\documentclass[journal]{IEEEtran}

\ifCLASSOPTIONcompsoc
  \usepackage[nocompress]{cite}
\else
  \usepackage{cite}
\fi

\usepackage{times}
\usepackage{epsfig}
\usepackage{graphicx}
\usepackage{amsmath}
\usepackage{amssymb}
\usepackage{pdfpages}
\usepackage{caption}
\usepackage{subcaption}
\usepackage{xcolor}
\usepackage{hyperref}

\ifCLASSINFOpdf
\else
\fi

\begin{document}

\title{Learning Selective Sensor Fusion for \\ States Estimation}

\author{Changhao Chen*,
        Stefano Rosa,
        Chris Xiaoxuan Lu,
        Bing Wang,
        Niki Trigoni,
        Andrew Markham
\IEEEcompsocitemizethanks{
\IEEEcompsocthanksitem Changhao Chen is with the College of Intelligence Science and Technology, National University of Defense Technology, Changsha, 410073, China
\IEEEcompsocthanksitem Stefano Rosa is with Istituto Italiano di Tecnologia (IIT), Genoa, Italy
\IEEEcompsocthanksitem Chris Xiaoxuan Lu is with the School of Informatics, University of Edinburgh, Edinburgh, EH8 9AB, United Kingdom
\IEEEcompsocthanksitem Bing Wang, Niki Trigoni and Andrew Markham are with the Department of Computer Science, University of Oxford, Oxford OX1 3QD, United Kingdom
\IEEEcompsocthanksitem *Corresponding author: Changhao Chen (changhao.chen@cs.ox.ac.uk)
}
\thanks{This work was supported by EPSRC Program ``ACE-OPS: From Autonomy to Cognitive assistance in Emergency OPerationS" (Grant number: EP/S030832/1) and NFSC (Grant number: 62103427, 62073331)}
\thanks{{The code of this work is available at https://github.com/changhao-chen/selective\_sensor\_fusion}}
}

\markboth{IEEE Transactions on Neural Networks and Learning Systems,~Vol.~X, No.~X, May~2022}%
{Shell \MakeLowercase{\textit{et al.}}: Bare Demo of IEEEtran.cls for Computer Society Journals}

\maketitle

\begin{abstract}
Autonomous vehicles and mobile robotic systems are typically equipped with multiple sensors to provide redundancy. By integrating the observations from different sensors, these mobile agents are able to perceive the environment and estimate system states, e.g. locations and orientations. Although deep learning approaches for multimodal odometry estimation and localization have gained traction, they rarely focus on the issue of robust sensor fusion - a necessary consideration to deal with noisy or incomplete sensor observations in the real world. Moreover, current deep odometry models suffer from a lack of interpretability. To this extent, we propose SelectFusion, an end-to-end selective sensor fusion module which can be applied to useful pairs of sensor modalities such as monocular images and inertial measurements, depth images and LIDAR point clouds. Our model is a uniform framework that is not restricted to specific modality or task. During prediction, the network is able to assess the reliability of the latent features from different sensor modalities and estimate trajectory both at scale and global pose. In particular, we propose two fusion modules - a deterministic soft fusion and a stochastic hard fusion, and offer a comprehensive study of the new strategies compared to trivial direct fusion. We extensively evaluate all fusion strategies in both public datasets  and on progressively degraded datasets that present synthetic occlusions, noisy and missing data and time misalignment between sensors, and we investigate the effectiveness of the different fusion strategies in attending the most reliable features, which in itself, provides insights into the operation of the various models.

\end{abstract}


\section{Introduction}
Mobile agents are often outfitted with multiple sensors. For example, a self-driving vehicle is equipped with a combination of GPS, IMUs, cameras,  {and} LIDAR.
Making such mobile agents fully autonomous and intelligent requires the ability of  {sensor} fusion, a method that can effectively exploit the individual strengths of distinct sensors and coherently estimate the system states. 
Multimodal sensor fusion has long been a central problem in robotics and computer vision \cite{thrun2005probabilistic}, with applications to perception, planning and control. 
Despite different application scenarios, the rationale for sensor fusion is more or less the same: many system state variables are not always fully observable by a single sensor modality, and combining different sensors that are complementary to each other can reduce the overall uncertainty, improving accuracy and/or robustness. 
Conventional sensor fusion methods resort to handcrafted design that heavily relies on human experience and domain knowledge. Consequently, the developed fusion methods are often modality-specific and/or task-specific. 

    \begin{figure}
      \centering
         \includegraphics[width=0.5\textwidth]{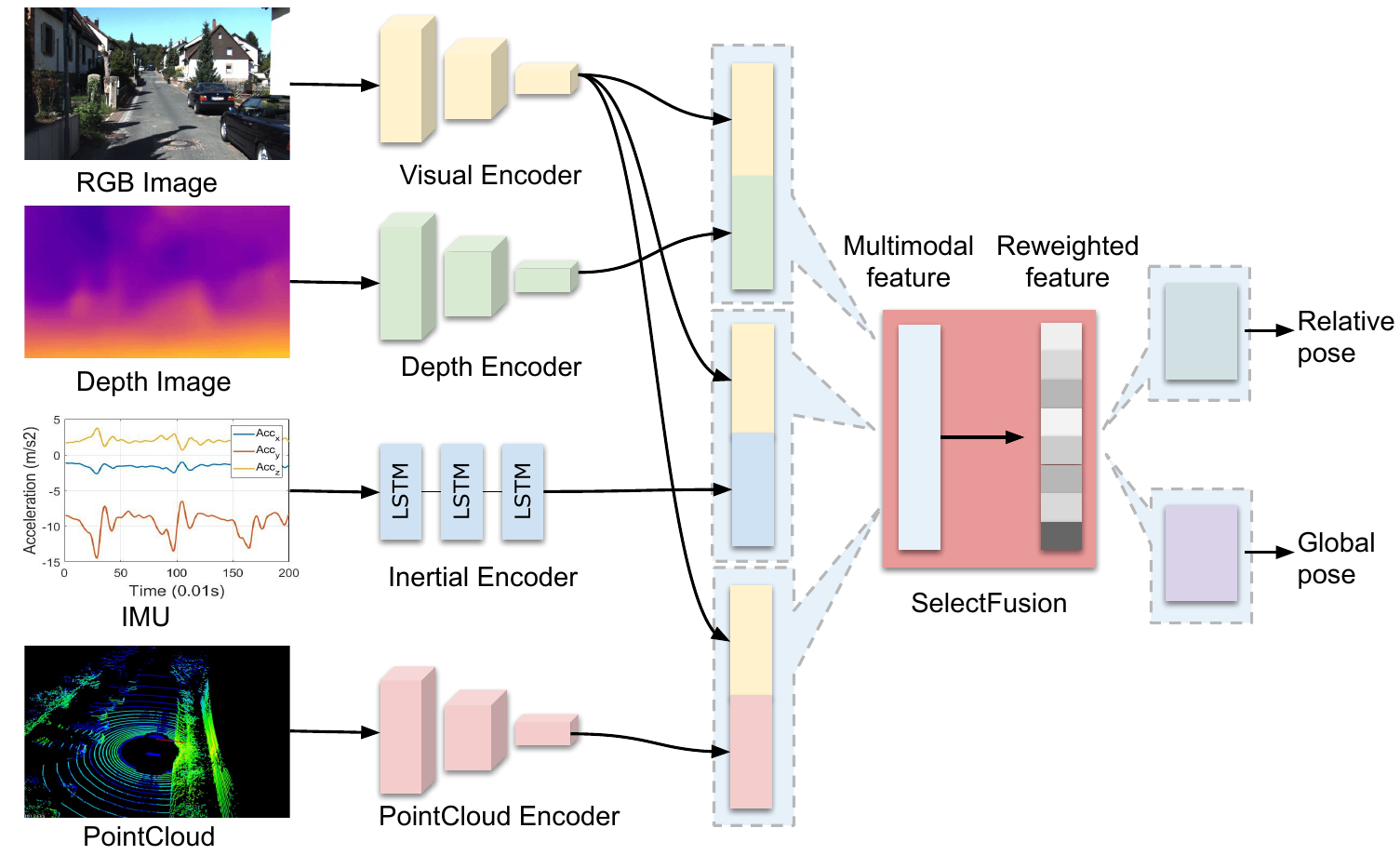}
         \caption{ An overview of the general framework to learn system states from multiple sensor modalities. Our framework can selectively utilize the suitable features for solving problems to improve both the accuracy and robustness.
         }
         \label{fig:overview}
     \end{figure}

Recently, there  {are} growing  {interests} in applying deep neural networks (DNNs) for \textit{learning to estimate system states} in an end-to-end manner, for example, solving Visual Odometry (VO) \cite{deepvo,zhou2017unsupervised,kashyap2020sparse}, Visual-Inertial Odometry (VIO) \cite{clarkwang2017,shamwell2019unsupervised} or camera relocalization \cite{kendall2015posenet,Clark2017}. Instead of building analytical models by hand, they are achieved by learning complex  {mapping functions} directly from raw sensory data to target values.
These end-to-end approaches are appealing due to the capability of deep networks to automatically extract features from high-dimensional raw data. However, despite the long history of classical sensor fusion techniques, there is a lack of effective fusion strategies working on the deep feature space, especially for the tasks of localization and odometry estimation. 
These previous learning-based methods are not explicitly modelling error sources in real-world usage. 
Without considering possible sensor errors, all features are directly fed into other modules for further pose regression in \cite{Brahmbhatt2018,Clark2017,kendall2015posenet}, or simply concatenated as in \cite{clarkwang2017}. These factors can possibly cause troubles to the accuracy and safety of neural systems, when input data are corrupted or missing. Moreover, the features from different modalities are considered equally important in these methods, although the complementary property of different modalities requires systems to utilise deep features with regard to observation uncertainties or self/environmental dynamics.

For this reason, we present a generic framework that models feature selection for robust sensor fusion, as illustrated in Figure \ref{fig:overview}. This work mainly considers the problem of using a pair of sensor modalities, although it can be extended naturally to three or more modalities. As a case study, two tasks - learning global localization and ego-motion estimation, are chosen to demonstrate the effectiveness of our proposed selective sensor fusion. 
Our system is not restricted to specific modality, performing feature selection from four different sensor data, i.e. RGB-images, inertial measurements, LIDAR point clouds and depth images.
The selection process is conditioned on the measurement reliability and the dynamics of both self-motion and environment. Two alternative feature weighting strategies are presented: soft fusion, implemented in a deterministic fashion; and hard fusion, which introduces stochastic noise and intuitively learns to keep the most relevant feature representations, while discarding useless or misleading information. 

By explicitly modelling the selection process, we are able to demonstrate the strong correlation between selected features and  environmental/measurement dynamics by visualizing the sensor fusion masks, as illustrated in Figure \ref{fig:mask}. In the case of estimating visual-inertial odometry, our results show that features extracted from different modalities (i.e., vision and inertial motion) are complementary in various conditions:  inertial features contribute more in presence of fast rotation, while visual features are preferred during large translations (Figure \ref{fig:correlation}). Thus, the selective sensor fusion provides insights into the underlying strengths of each sensor modality, guiding future multimodal system design. We demonstrate how incorporating selective sensor fusion makes neural models robust to data corruption typically encountered in real-world scenarios. 

This paper builds on the work published in \cite{chen2019selective},
and presents a generic framework for selective sensor fusion in multimodal deep pose estimation. 
 {This} work 
extends the fusion strategies from visual-inertial odometry to the  {problem} of  {learning} LIDAR-visual odometry and RGB-depth relocalization.
To summarise, the novel contributions of this work are as follows:
\begin{itemize}
    \item We present  {SelectFusion}, a novel generic framework to learn selective sensor fusion enabling more robust and accurate odometry and localization in real-world scenarios.
    \item We show how our selective sensor fusion can be incorporated into a uniform framework, not restricted by specific modality or task, by learning odometry estimation or relocalization on fusing a pair of modalities from vision, depth, inertial and LIDAR data.
    \item Our  {SelectFusion} masks can be visualized and interpreted, providing deeper  {insights} into the relative strengths of each stream, and guiding  {future} system design.
    \item We create challenging datasets on top of current public datasets by considering seven different sources of sensor degradation, and conduct a new and complete study on the accuracy and robustness of deep sensor fusion in presence of corrupted data.
\end{itemize}

The reminder of the paper is organized as follows: Section \ref{sec: related work} contains a survey of related work; Section \ref{sec: framework} presents a generic framework for multimodal sensor fusion; Section \ref{sec: selective sensor fusion} introduces our proposed selective sensor fusion mechanism; Section \ref{sec: experiments} evaluates SelectFusion applied to three multimodal models for relocalization and trajectory estimation through extensive experiments; Section \ref{sec: conclusion} finally draws conclusions.


\section{Background and Related Work}
\label{sec: related work}
\subsection{Learning-based Pose Estimation} 
\label{sec: learning localization}
\noindent\textbf{Visual-inertial Odometry:}
Recent work  {shows} how it is possible to learn to estimate odometry from inertial data using recurrent neural networks \cite{ionet2018}, making deep visual-inertial odometry estimation possible. 
VINet \cite{clarkwang2017}  {uses} neural network to learn visual-inertial odometry, by directly concatenating visual and inertial features. 
VIOLearner \cite{shamwell2019unsupervised} presents an online error correction module for deep visual-inertial odometry that estimates the trajectory by fusing RGB-D images with inertial data.
DeepVIO \cite{han2019deepvio} recently {proposes} a fusion network to fuse visual and inertial features. This network is trained with a dedicated loss. However, this way of learning sensor fusion does not expose the behaviour of the fusion module, while we propose the use of an interpretable mask, that offers insight into the usefulness of the input at any time.
We  {observe} that previous methods do not properly address the problem of learning a meaningful sensor fusion strategy, but simply concatenate visual and inertial features in  latent space.

\noindent\textbf{LIDAR Odometry:}
Learning LIDAR odometry has been explored by LO-Net \cite{li2019net}, which exploits geometric consistency for scan-to-scan motion estimation, while also learning pose correction similarly to deep SLAM approaches, and can achieve accuracy comparable to traditional approaches \cite{lu2019l3}. 
Fusion of LIDAR and visual information has been investigated in \cite{graeter2018limo}, which proposes to fuse LIDAR and visual information, but in their work the learning is limited to training a model for removal of moving objects rather than localization.

\noindent\textbf{Camera Relocalization}
Deep approaches have also been devoted to visual localization. Posenet\cite{kendall2015posenet}  {is} the first work to use Convolutional Neural Networks (CNNs) for 6-DoF pose regression from monocular images. PoseNet has been further improved by combining CNNs and LSTMs for feature correlation\cite{walch2017image}, introducing temporal information\cite{Clark2017}, incorporating spatial constraints\cite{Brahmbhatt2018} or by adding additional co-visibility constraints based on local maps and the estimated odometry \cite{xue2019local}. {MS-Transformer \cite{shavit2021learning} is a recent relocalization work based on transformer architecture, achieving the state-of-the-art results. }

\subsection{Multimodal Learning} 
Multimodal learning aims to solve machine learning problems involving multiple data modalities.  {They are generally categorized into aggregation-based and alignment-based fusion methods.}
The success of multimodal learning has been demonstrated in a wide range of applications, e.g. video captioning \cite{song2018deterministic}, medical image retrieval \cite{gu2020deep}, face recognition \cite{ding2015robust}, manipulation \cite{lee2019making}, autonomous navigation \cite{liu2017learning} and  {body-sensor-networks \cite{wang2020multi}}.
 {Recently, \cite{wang2020deep} designs a Channel-Exchanging-Network (CEN) that fuses multiple modalities by dynamically exchanging the channels of sub-networks. \cite{li2021rfn} proposes a residual fusion network (RFN) based framework that automatically extracts features and fuses multi-scale features. In \cite{shu2015weakly,tang2016generalized}, a shared layer is designed to transfer cross-modal features, in which an inner product function of extracted features from two modalities is adopted to combine them for domain transferring. This inner product method is similar to our soft fusion but for different purposes.
}
However, there is a lack of systematic study into the sensor fusion for deep state estimation, especially in learning based localization and pose estimation, as discussed in Section \ref{sec: learning localization}. 

\subsection{Attention Mechanism}
Our proposed selective sensor fusion is particularly related to  attention mechanisms, that have been widely applied in neural machine translation \cite{galassi2020attention}, image caption generation \cite{Xu2015}, visual question answering \cite{liu2020adversarial} and video description \cite{Hori2017}. 
Limited by the fixed-length vector in embedding space, these attention mechanisms compute a focus map to help the decoder, when generating a sequence of words. 
This is different from our design intention that the features selection works to fuse multimodal sensor fusion for deep pose estimation, and cope with more complex error resources, and self-motion dynamics.

    \begin{figure*}
      \centering
         \includegraphics[width=0.85\textwidth]{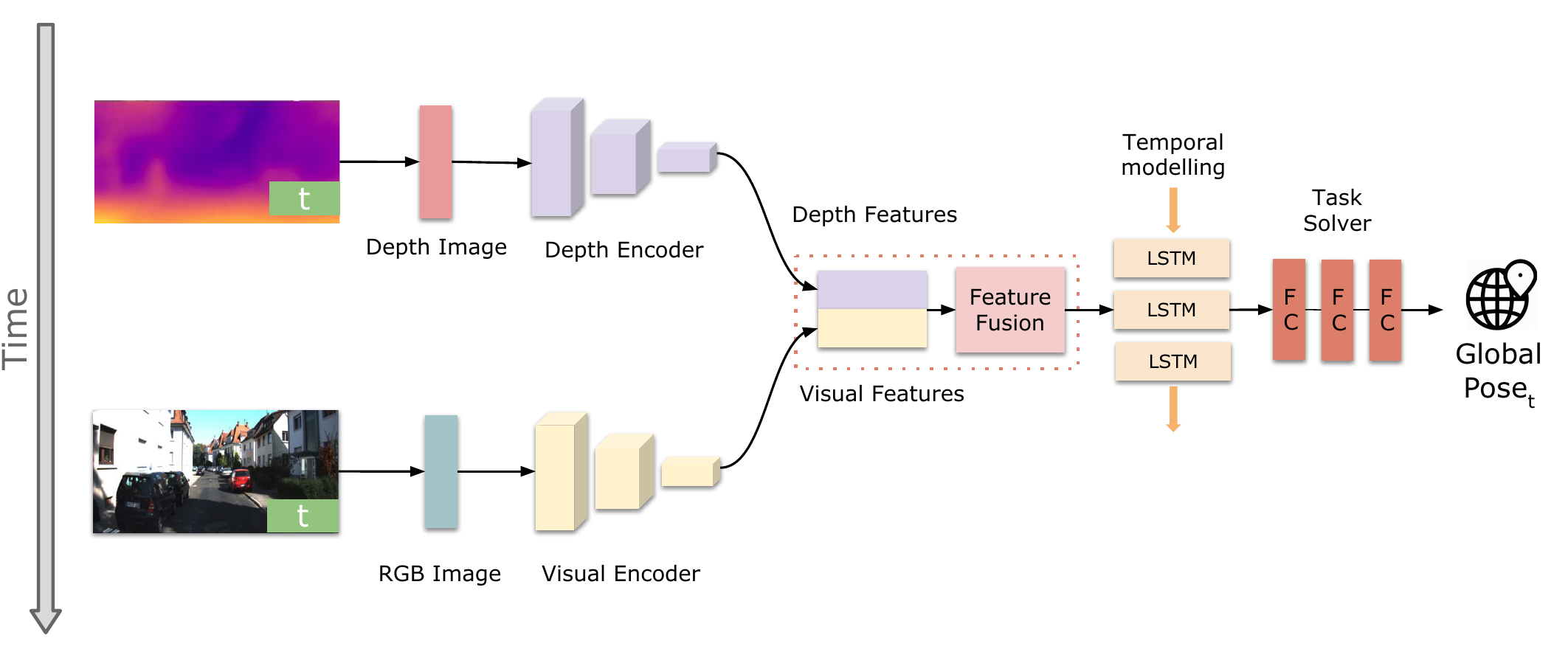}
         \caption{An overview of our depth-vision relocalization (\textbf{Task 1}) architecture with proposed selective sensor fusion, consisting of depth and visual encoders, feature fusion, temporal modelling and task solver (global pose estimation). 
         }
         \label{fig: depth-vision framework}
     \end{figure*}
     
     \begin{figure*}
      \centering
         \includegraphics[width=0.85\textwidth]{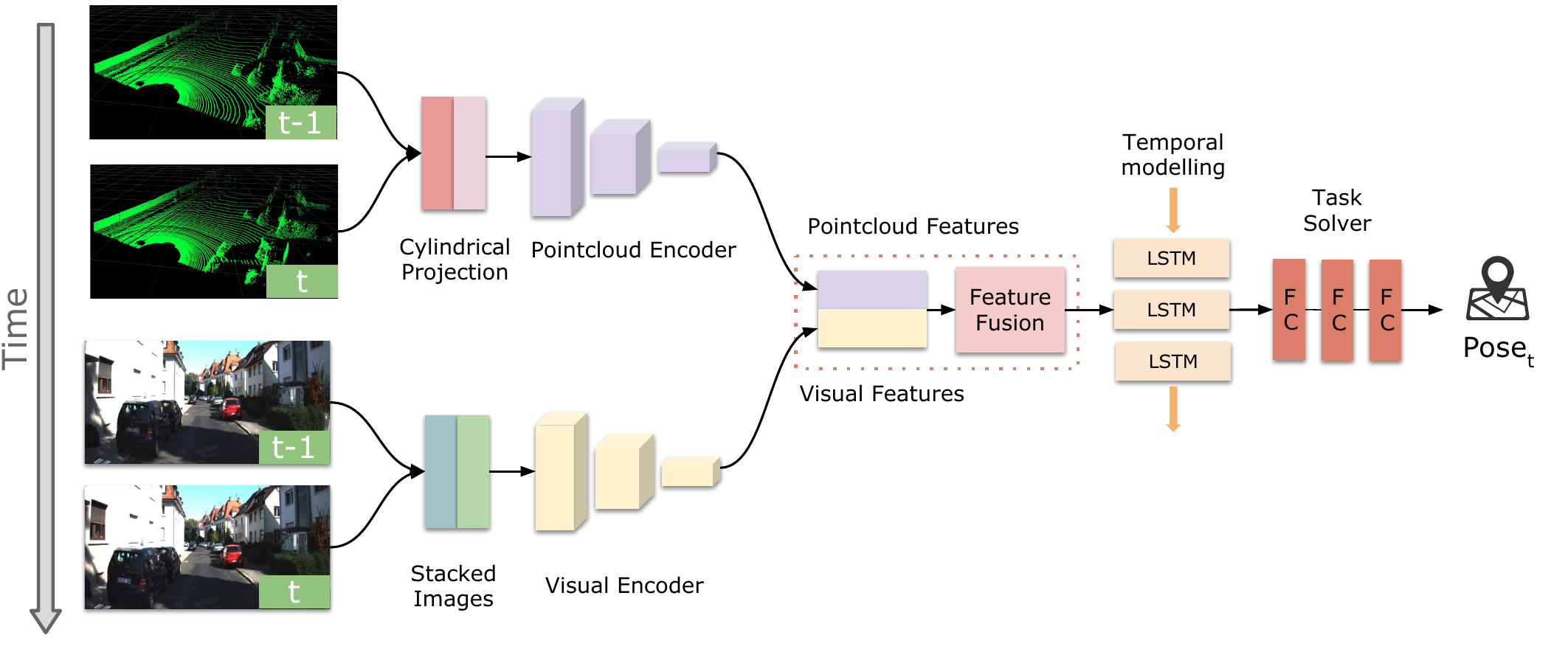}
         \caption{An overview of our neural LIDAR-visual odometry (\textbf{Task 2}) architecture with proposed selective sensor fusion, consisting of visual and LIDAR encoders, feature fusion, temporal modelling and task solver (relative pose regression). 
         }
         \label{fig: LIDAR-vision framework}
     \end{figure*}

    \begin{figure*}
      \centering
         \includegraphics[width=0.85\textwidth]{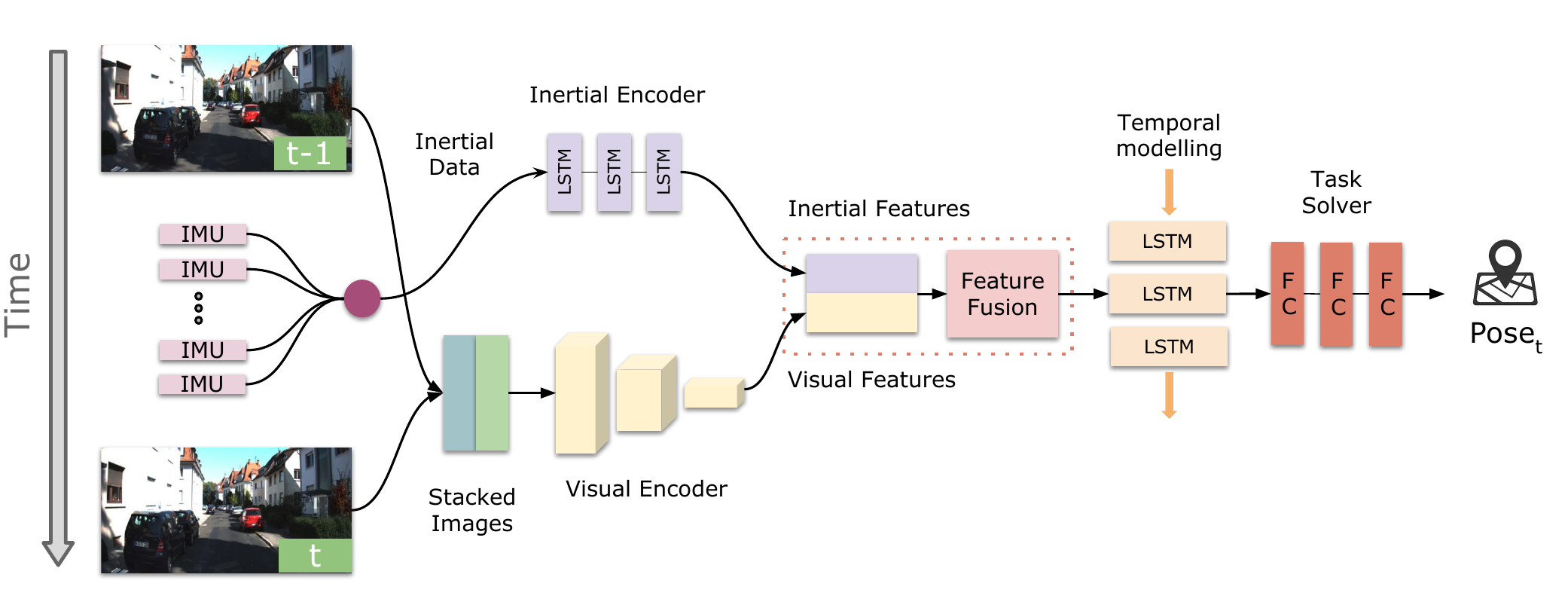}
         \caption{An overview of our neural visual-inertial odometry (\textbf{Task 3}) architecture with proposed selective sensor fusion, consisting of visual and inertial encoders, feature fusion, temporal modelling and task solver (relative pose regression). 
         }
         \label{fig: visual-inertial framework}
     \end{figure*}

\section{Learning Multimodal Representations}
\label{sec: framework}
This section presents a uniform framework to learn multimodal representation for state estimation, which lays the foundation for selective sensor fusion. 
Figure \ref{fig: depth-vision framework}, \ref{fig: LIDAR-vision framework} and \ref{fig: visual-inertial framework}, show a modular overview of the architecture, consisting of feature encoders, feature fusion, temporal modelling and task solver. 

\subsection{Feature Encoders}
\subsubsection{Visual Feature Encoders}
As visual feature encoders are used in both global relocalization and odometry estimation, they are designed with respect to the property of each task for better feature extraction and utilization. 

For a relative pose (odometry) estimation, latent representations are extracted from a set of two consecutive monocular images $\mathbf{x}_V$. 
Ideally, we want our visual encoder $f_{\text{vision}}$ to learn geometrically meaningful features rather than features overfitted with appearance or context. 
For this reason, instead of using a PoseNet model \cite{kendall2015posenet}, as commonly found in other DL-based VO approaches \cite{zhou2017unsupervised,Zhan2018,Yin2018}, we use a FlowNet-style architecture, i.e. FlowNetSimple \cite{Fischer2015} as our feature encoder.
Flownet provides features that are suited for optical flow prediction, which highly contributes to the  {ego-motion} detection. 
The network consists of nine convolutional layers. The size of the receptive fields gradually reduces from 7$\times$7 to 5$\times$5 and finally 3$\times$3, with stride two for the first six  {layers}.
Each layer is followed by a ReLU nonlinearity except for the last one, and we use the features from the last convolutional layer
 $\mathbf{a}_V$ as our visual feature. We initialize the visual encoder with the weights of a model that was pre-trained on the FlyingChairs dataset\footnote{https://lmb.informatik.uni-freiburg.de/resources/datasets/FlyingChairs.en.html}, since training from scratch would require  {a} larger  {amount} of data compared with our dataset size. 

For a global relocalization task, we instead use Residual Neural Network (ResNet) \cite{he2016deep} to extract features from a set of single images. Both structure and appearance features contribute to the retrieval of absolute poses in the 3D scene that has been visited before. Hence, visual features should capture the entire scene. 
We adopt ResNet18, consisting of 18 layers convolutional layers with skip connections, and modify it by introducing an average pooling layer and a full-connected layer at the end, that  {transform} the features after ResNet18 to a $d$ dimension visual feature $\mathbf{a}_V$.

In summary, given a set of images $\mathbf{x}_V$, we are able to extract visual features $\mathbf{a}_V  \in \mathbb{R}^d$  {suited to} the task via the Visual Encoder (FlowNet) or (ResNet) $f_{\text{vision}}$: 
\begin{equation}
    \mathbf{a}_V = f_{\text{vision}}(\mathbf{x}_V).
\end{equation}

\subsubsection{Inertial Feature Encoder}
  
Inertial data streams have a strong temporal component, and are generally available at higher frequency ($\sim$100 Hz) than images ($\sim$10 Hz). In order to model the temporal dependencies of consecutive inertial measurements, we use a two-layer Bi-directional LSTM with 128 hidden states as the Inertial Feature Encoder $f_{\text{inertial}}$. In the deep VIO model, as shown in Figure \ref{fig: visual-inertial framework}, a window of inertial measurements $\mathbf{x}_I$ between each two images is fed to the inertial feature encoder in order to extract the $d$ dimensional feature vector ${\mathbf{a}_I}  \in \mathbb{R}^d$:

\begin{equation}
    \mathbf{a}_I = f_\text{inertial}(\mathbf{x}_I).
\end{equation}

\subsubsection{Depth Feature Encoder}

In our work, depth image is exploited to solve the task of vision-depth based relocalization, as shown in Figure \ref{fig: depth-vision framework}.
Similar to the visual encoder designed for relocalization, we also use ResNet18 as depth feature encoder, but replace the first layer of ResNet model with a 1-channel convolutional network, considering that depth image is 1-channel rather than 3-channels. 
Hence, the input is a set of 1-channel depth images $\mathbf{x}_D$, and transformed into a $d$ dimensional features vector $\mathbf{a}_D  \in \mathbb{R}^d$ via the depth encoder $f_\text{depth}$:
\begin{equation}
    \mathbf{a}_D = f_\text{depth}(\mathbf{x}_D).
\end{equation}

\subsubsection{Pointcloud Feature Encoder}
The point clouds are a set of data in Cartesian coordinates, representing 3D structure in space. They are produced normally by LIDAR devices. The sparse structure and irregular format of point cloud data make them hard to be processed directly by neural networks. To allow convolutional neural networks (CNNs) to effectively process point cloud data, we convert them into a regular point cloud matrix via the cylindrical projection \cite{chen2017multi,li2019net}:
\begin{equation}
    \alpha = \text{arctan} (y/x)/\Delta \alpha
\end{equation}
\begin{equation}
    \beta = \text{arcsin} (z/\sqrt{x^2 + y^2 + z^2})/\Delta \beta
\end{equation}
where (x, y, z) are original coordinates in LIDAR coordinate system, and ($\alpha$, $\beta$) are new coordinates in the point cloud matrix. The new point cloud matrix is with a size of $H \times W \times C$. 
The position ($\alpha$, $\beta$) of matrix is filled with the range value $r=\sqrt{x^2 + y^2 + z^2}$ from the position (x, y, z) of original point cloud.

In this work, point cloud data are used to learn vision-LIDAR odometry, as shown in Figure \ref{fig: LIDAR-vision framework} and hence we also use the FlowNet visual encoder to transform the input matrix $\mathbf{x}_P$ into a $d$ dimensional point cloud feature $\mathbf{a}_P \in \mathbb{R}^d$:

\begin{equation}
    \mathbf{a}_P = f_\text{pointcloud}(\mathbf{x}_P).
\end{equation}

    \begin{figure*}
      \centering
         \includegraphics[width=0.90\textwidth]{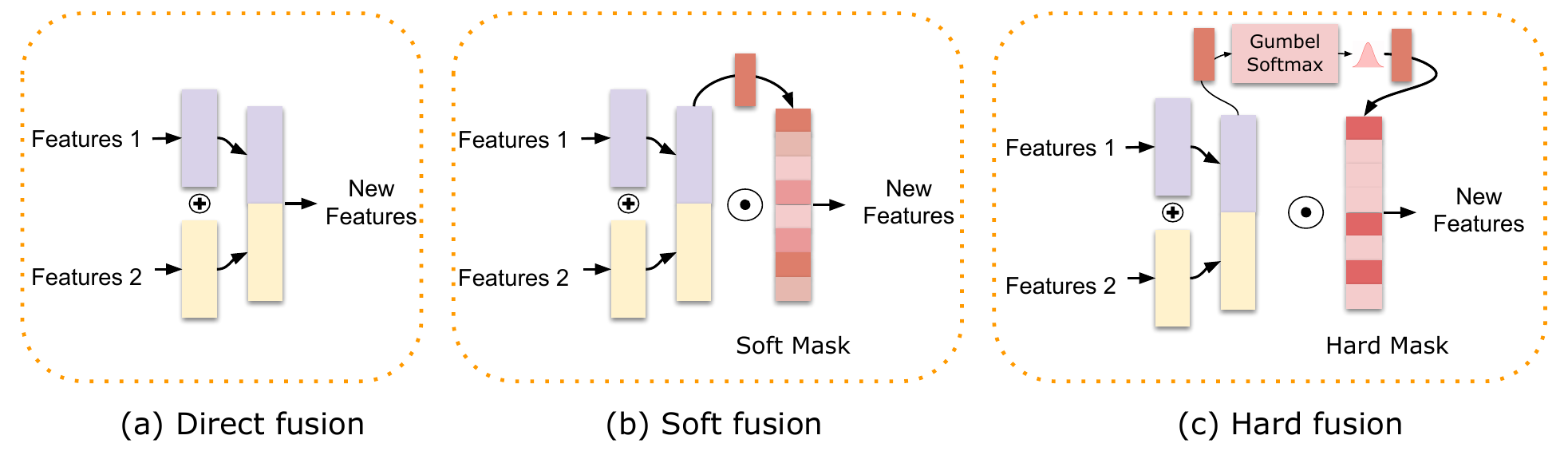}
         \caption{An overview of three fusion methods: (a) direct fusion, (b) soft fusion and (c) hard fusion.
         }
         \label{fig: fusion overview}
     \end{figure*}

\subsection{Fusion Function}
We now combine the high-level representation produced by each feature encoder from raw data sequences, with a fusion function $g$ that combines information from a pair of sensor modalities to extract the useful combined feature $\mathbf{z}$ for a regression task:
    \begin{equation}
        \mathbf{z} = g(\mathbf{a}_1, \mathbf{a}_2),
    \end{equation}
where $(\mathbf{a}_1, \mathbf{a}_2)$ is any pair of sensor modality features from visual $\mathbf{a}_V$, inertial $\mathbf{a}_I$, depth $\mathbf{a}_D$, and point cloud $\mathbf{a}_P$ channels. In this work, we specifically investigate the problem of fusing two sensor modalities for better demonstration on existing datasets, although our framework can extend naturally to exploit three or more modalities.

There are several different ways to implement this fusion function.
The current approach is to directly concatenate the two features together into one feature space (we call this method direct fusion $g_\text{direct}$). However, in order to learn a robust sensor fusion model, we propose two fusion schemes -- deterministic soft fusion $g_\text{soft}$ and stochastic hard fusion $g_\text{hard}$, which explicitly model the feature selection process according to the current environment dynamics and the reliability of the data input.  
The fusion network is another deep neural network module. Details will be discussed in Section~\ref{sec: selective sensor fusion}.

\subsection{Temporal Modelling and Task Solvers}

The fundamental tenet of state estimation requires modelling temporal dependencies to derive accurate system states, e.g. relative poses. In the past, a state-space-model (SSM) describes this temporal relation and evolution of system states.
Similarly, in our learning model, a recurrent neural network, i.e. Long Short-Term Memory (LSTM) network takes in the input combined feature representation $\mathbf{z}_{t}$ at time step $t$ and its previous hidden states $\mathbf{h}_{t-1}$ and models the dynamics and connections between a sequence of features. The hidden states $\mathbf{h}_{t}$  {contain} the history of the features relevant to the task.
After the recurrent network, a fully-connected layer serves as the regressor,  {transforming} the features to a system state $\mathbf{y}_t$, i.e. pose transformation or global pose, representing the motion transformation over a time window or a global location/orientation. 

Hence, the relation between the final system states $\mathbf{y}_t$ and the input features $\mathbf{z}_{t}$ can be described via a recurrent neural network and previous hidden states $\mathbf{h}_{t-1}$:
    \begin{equation}
        \mathbf{y}_t = \mathbf{RNN}(\mathbf{z}_{t}, \mathbf{h}_{t-1}) .
    \end{equation}
We implemented three tasks based on this multimodal representation learning framework to estimate key system states from pairs of raw sensory data.

\subsubsection{Task 1: Learning Vision-Depth Relocalization}

The first task is to exploit monocular RGB images and depth images to perform global relocalization in the scenarios that have been visited before. As illustrated in Figure \ref{fig: depth-vision framework}, depth and RGB images are encoded into features by the Depth Encoder and Visual Encoder (ResNet), fused as new features through Feature Fusion modules, and converted into global poses via temporal modelling and task regression modules. The global pose $\mathbf{y}=[\mathbf{p}, \mathbf{q}]$ is composed by a 3-D position vector $\mathbf{p} \in \mathbb{R}^3$ and a quaternion $\mathbf{q} \in \mathbb{R}^4$ for orientation. The objective is to minimize the L1 distance between the groundtruth values $[\mathbf{\hat{p}}, \mathbf{\hat{q}}]$ and predicted values $[\mathbf{p}, \mathbf{q}]$ with the loss function: 
\begin{equation}
    L(\theta)_1 = |\mathbf{\hat{p}}-\mathbf{p}| + \lambda_1 \begin{vmatrix}\mathbf{\hat{q}} - \mathbf{\frac{q}{||q||}}\end{vmatrix},
\end{equation}
where $\lambda_1$ is a balance factor, which we choose as $\lambda_1=10$ in our experiment. Here, L1 loss is chosen rather than L2 loss, because L1 loss performs better and more stable \cite{kendall2017geometric}.  

\subsubsection{Task 2: Learning Lidar-Vision Odometry}

The second task is to learn LIDAR-vision odometry. Different from global relocalization, odometry estimation produces relative poses between two frames of images, which can adapt to new scenarios. Global pose is achieved by integrating pose transformations. As shown in Figure \ref{fig: LIDAR-vision framework}, the framework consists of Point Cloud Encoder and Visual Encoder (FlowNet) that extract features from LIDAR point cloud data and RGB images, Feature Fusion that combines LIDAR and visual features as a new feature vector, and Temporal Modelling and Task Solver modules to transform features as system states. The network outputs relative poses $\mathbf{y}=[\mathbf{p}, \mathbf{r}]$, consisting of a 3-dimensional translation vector $\mathbf{p} \in \mathbb{R}^3$, and a 3-dimensional Euler rotation vector $\mathbf{r} \in \mathbb{R}^3$. The objective is to minimize the Mean Square Error (MSE) of the relative poses to recover optimal neural networks parameters $\theta$:
\begin{equation}
    L(\theta)_2 = ||\mathbf{\hat{p}}-\mathbf{p}||_2 + \lambda_2 ||\mathbf{\hat{r}} - \mathbf{r}||_2,
\end{equation}
where $[\mathbf{\hat{p}}, \mathbf{\hat{r}}]$ are groundtruth values, and $\lambda_2$ is a scale factor to balance between translational error and rotational error. $\lambda_2$ is chosen as $100$ in our experiment.

\subsubsection{Task 3: Learning Visual-Inertial Odometry}
The third task is to learn visual-inertial odometry, providing accurate pose estimation by using visual and inertial sensors, which are widely deployed in mobile robotics, self-driving vehicles and drones. Similar to LIDAR-vision odometry, our model outputs the relative poses between two frames of images.  
Figure \ref{fig: visual-inertial framework} shows that visual and inertial features are extracted from consecutive monocular images, and a sequence of inertial data between two frames of images by FlowNet based Visual Encoder and LSTM based Inertial Encoder. The features are combined as new features via Feature Fusion, and converted into system states through Temporal Modelling and Task Regressor. The network produces pose transformation $\mathbf{y}=[\mathbf{p}, \mathbf{r}]$ with a 3-dimensional translation vector $\mathbf{p} \in \mathbb{R}^3$, and a 3-dimensional rotation vector $\mathbf{r} \in \mathbb{R}^3$ (the rotation vector is represented by 3-dimensional Euler angles). By minimizing the MSE of the predicted relative poses, the optimal parameters $\theta$ are recovered via:
\begin{equation}
    L(\theta)_3 = ||\mathbf{\hat{p}}-\mathbf{p}||_2 + \lambda_3 ||\mathbf{\hat{r}} - \mathbf{\hat{r}}||_2,
\end{equation}
where $[\mathbf{\hat{p}}, \mathbf{\hat{r}}]$ are true relative poses, $[\mathbf{p}, \mathbf{r}]$ are predicted values, and $\lambda_3$ is a scale factor to balance between translational error and rotational error. In our case, we choose $\lambda_3$ as $100$.

\section{Selective Sensor Fusion}
\label{sec: selective sensor fusion}
In this section, we propose SelectFusion, a generic framework to selectively learn multisensory representation from raw data.
Intuitively, the features from each modality offer different strengths for the task of state estimation. 
Our perspective is that simply considering all features as that they are equally important and correct, without any consideration of degradation and self/environmental dynamics, is unwise and will lead to unrecoverable drifts and errors. Therefore, we propose two different selective sensor fusion schemes for explicitly learning the feature selection process: soft (deterministic) fusion, and hard (stochastic) fusion, as illustrated in Figure \ref{fig:feature_fusion}. In addition, we also present a straightforward sensor fusion scheme -- direct fusion -- as a baseline model for comparison.

\subsection{Direct Fusion}
A straightforward approach for implementing sensor fusion consists in the use of Multi-Layer Perceptrons (MLPs) to combine the features from the two sensor modality channels. 
Ideally, the system learns to discriminate relevant features for prediction in an end-to-end fashion.
Hence, direct fusion is modelled as:
\begin{equation}
    g_\text{direct}(\mathbf{a}_1, \mathbf{a}_2) = [\mathbf{a}_1; \mathbf{a}_2]
\end{equation}
where $[\mathbf{a}_1; \mathbf{a}_2]$ denotes an operation function that concatenates features $\mathbf{a}_1$ and $\mathbf{a}_2$, which are extracted from the Modality One and Two respectively.

\subsection{Soft Fusion (Deterministic)}

We now propose a soft fusion scheme that explicitly and deterministically models feature selection.
Similar to the widely applied attention mechanism \cite{Vaswani2017,Xu2015,Hori2017}, this function re-weights each feature by conditioning on both  sensor modality channels, allowing the feature selection process to be jointly trained with other modules. The function is deterministic and differentiable.

Here, a pair of continuous masks $\mathbf{s_1}$ and $\mathbf{s_2}$  {are} introduced to implement soft selection of the extracted feature representations, before these features are passed to temporal modelling and task solver:
    \begin{eqnarray}
        &\mathbf{s}_1 = \text{Sigmoid}(\text{MLP}_1([\mathbf{a}_1; \mathbf{a}_2])) \\
        &\mathbf{s}_2 = \text{Sigmoid}(\text{MLP}_2([\mathbf{a}_1; \mathbf{a}_2]))
    \end{eqnarray}
where $[\mathbf{a}_1; \mathbf{a}_2]$ denotes an operation function that concatenates features $\mathbf{a}_1$ and $\mathbf{a}_2$. $\text{MLP}$ is multilayer perceptron, a feedforward neural network that  {transforms} features to fusion mask space. The Sigmoid function makes sure that each of the features will be re-weighted in the range $[0,1]$. This process is deterministically parameterised by the neural networks, conditioned on both the features $\mathbf{a}_1$ and features $\mathbf{a}_2$.
$\mathbf{s_1}$ and $\mathbf{s_2}$ represent soft masks applied to the features extracted from Modality One and Modality Two respectively. 

Then, the visual and inertial features are element-wise multiplied with their corresponding soft masks as the new re-weighted vectors. The selective soft fusion function is modelled as
\begin{equation}
    g_{\text{soft}}(\mathbf{a}_1, \mathbf{a}_2) = [\mathbf{a}_1 \odot \mathbf{s}_1; \mathbf{a}_2 \odot \mathbf{s}_2].
\end{equation}

\subsection{Hard Fusion (Stochastic)}

    \begin{figure}[t]
      \centering
         \includegraphics[width=0.95\columnwidth]{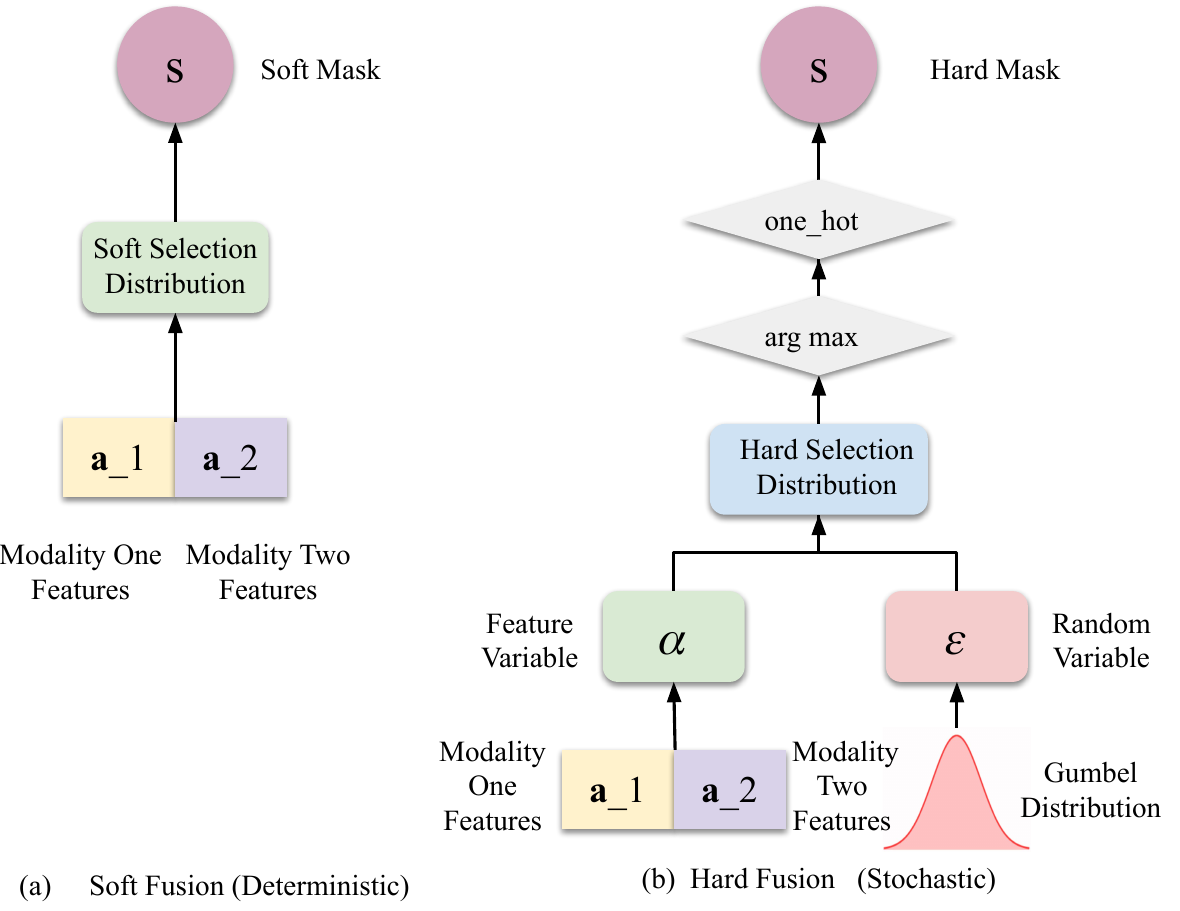}
         \caption{An illustration of our proposed soft (deterministic) and hard (stochastic) feature selection process.}
         \label{fig:feature_fusion}
     \end{figure}
 
In addition to the soft fusion introduced above, we propose a variant of the fusion scheme -- hard fusion.
Instead of re-weighting each feature with a continuous value, hard fusion learns a stochastic function that generates a binary mask that either propagates the feature or blocks it.
This mechanism can be viewed as a switcher for each component of the feature map, which is a stochastic layer implemented by a parametrised Bernoulli distributions. 

However, the stochastic layer cannot be trained directly by back-propagation, as gradients will not propagate through discrete latent variables. 
To tackle this, the REINFORCE algorithm \cite{williams1992simple,mnih2014neural} is generally used to construct the gradient estimator.
In our case, we propose to employ a more lightweight method -- Gumbel-Softmax resampling \cite{jang2016categorical,maddison2016concrete} to infer the stochastic layer of hard fusion, so that our hard fusion module can be trained in an end-to-end fashion as well.

 {Before training the model, the distribution of hard mask is unknown. As each element of this mask $\mathbf{s}^{(i)}$ is a single stochastic variable, we assume it to be under Bernoulli distribution, which takes the value 1 with probability $p$ and the value 0 with probability $q=1-p$. As the entire mask $\mathbf{s}$ consists of $n$ elements, it is under the binomial distribution with parameters $n$, which is the discrete probability distribution of the number of successes in a sequence of $n$ independent experiments.}
 {Therefore,} instead of learning masks deterministically from features, hard masks $\mathbf{s}_1$ and $\mathbf{s}_2$, representing the binary mask for the features from two modalities, are re-sampled from a discrete Binomial distribution. This discrete distribution is parameterized by $\pmb{\alpha}$, which is learned by deep neural networks and conditioned on features but with the addition of stochastic noise:
    \begin{eqnarray}
       &\mathbf{s}_1 \sim  p(\mathbf{s}_1 | \mathbf{a}_1,\mathbf{a}_2) = \text{Binomial}(\pmb{\alpha}) \\
       &\mathbf{s}_2 \sim p(\mathbf{s}_2 | \mathbf{a}_1,\mathbf{a}_2) = \text{Binomial}(\pmb{\alpha}),
    \end{eqnarray}
where each mask $\mathbf{s} = [\mathbf{s}^{(1)}, ..., \mathbf{s}^{(n)}]$ is a n-dimensional binary vector $\mathbf{s}^{(i)}$. Each element of hard mask $\mathbf{s}^{(i)}$ is a 2-dimensional categorical variable, deciding whether to select the $i$th feature or not. The total number of features is $n$. The element $\mathbf{s}^{(i)}$ can be viewed as resampling from a Bernoulli distribution:
    \begin{equation}
        \mathbf{s}^{(i)} \sim \text{Bernoulli}(\pmb{\alpha}^{(i)}).
    \end{equation}

Similar to soft fusion, the features from two modalities are element-wise multiplied with their corresponding hard masks as new reweighted vectors. 
The stochastic hard fusion function is modelled as
    \begin{equation}
        g_\text{hard}(\mathbf{a}_1,\mathbf{a}_2)= [\mathbf{a}_1 \odot \mathbf{s}_1; \mathbf{a}_2 \odot \mathbf{s}_2]. 
    \end{equation}
    
Now we come to solve the problem of inferring this discrete distribution in order to generate hard mask $\mathbf{s}$.  
We apply the so-called Gumbel-Softmax trick to convert the non-continuous function into a continuous approximation by using the fact that the distribution of a discrete random variable $P(x=k)$ can be reparameterized by a random variable $\pi_k$ and a Gumbel random variable $\epsilon_k$ via
\begin{equation}
    x = \arg \max_{\substack{k}} (\log \pi_k + \epsilon_k).
\end{equation}

In practical, it is simple to implement this reparameterization trick into our model.
Figure \ref{fig:feature_fusion} (b) shows the detailed workflow of our proposed Gumbel-Softmax resampling based hard fusion. 
The Gumbel-max trick \cite{Maddison2014} allows us to efficiently draw a hard mask $\mathbf{s^{(i)}}$ from a categorical distribution given the class vector $\pi_{k}^{(i)}$ and a Gumbel random variable $\epsilon_{k}^{(i)}$, and then an one-hot encoding performs 'binarization' of the category:
    \begin{equation}
        \label{eq:s resample}
        \mathbf{s}^{(i)} = \text{one\_hot} (\arg \max_{\substack{k}} [\epsilon_{k}^{(i)}+\log \pi_{k}^{(i)}]),
    \end{equation} 
where $i \in [1,..,n]$ is the index of feature, $k \in [1, 2]$ is the index of class vector for each feature. In this case, there are only two categories, indicating whether to select a particular feature or not.
This can be viewed as a process of adding independent Gumbel perturbations to the discrete class variable. 
In practice, the random variable $\pmb{\epsilon}$ is sampled from a Gumbel distribution, which is a continuous distribution on the simplex that can approximate categorical samples:
\begin{equation}
    \epsilon = -\log(-\log(u)), u \sim \text{Uniform}(0,1).
\end{equation}
In Equation \ref{eq:s resample} the argmax operation is not differentiable, so softmax function is used as an approximation:
    \begin{equation}
        h^{(i)} = \frac{\exp((\log (\pi_{k}^{(i)}+\epsilon_{k}^{(i)})/\tau)}{\sum_{j=1}^{2} \exp((\log (\pi_{k}^{(i)})+\epsilon_{k}^{(i)})/\tau)}, k=1,2
    \end{equation}
where $\tau > 0$ is the temperature that modulates the re-sampling process. Finally, $h^{(i)}$ is transformed into a binary mask $\mathbf{s}^{(i)}$ through the one\_hot function. 

The $\pi_{k}^{(i)}$ is jointly learned by deep neural networks in our models, and formulated as the parameters $\pmb{\alpha}=(\pi_{k}^{i})|_{k=1,2}^{i=1..n}$ 
, conditioned on the concatenated feature vectors $[\bf{a}_1; \bf{a}_2]$ from two modalities:
    \begin{eqnarray}
       \pmb{\alpha} = \text{ReLU}(\text{FC}([\bf{a}_1; \bf{a}_2])),
    \end{eqnarray}
where $\text{FC}$ is full-connected layer, to map concatenated features into $k*2$ dimensional class vectors. $\text{ReLU}$ is to impose nonlinearity and ensures the class vectors to be nonnegative.

In our approach, we find that modulating the temperature with respect to the training procedure can enable better performance in selective sensor fusion. This is because the temperature determines the samples and gradients: when the temperature is high, the variance of the gradients is small, while the samples are more smooth; at low temperatures, the variance of the gradients is high, while the samples are more discrete, which means it will fit well into the discrete distribution of the fusion mask. Thus we start the temperature from a higher value, i.e. 1 in our case, and gradually decrease it towards 0.5 over each epoch of the training process.

\section{Experiments}
\label{sec: experiments}
We conducted extensive experiments above four well-known public datasets to learn from different pairs of sensor modalities: the 7-Scenes dataset \cite{Shotton2013} for vision-depth based relocalization (Task 1), the KITTI odometry dataset \cite{Geiger2013} for vision-LIDAR odometry estimation (Task 2), the KITTI raw dataset \cite{Geiger2013} and the EuRoC MAV dataset \cite{euroc} for visual-inertial odometry (Task 3).

\subsection{Experimental Setups}

Our frameworks were implemented with PyTorch and trained on a NVIDIA Titan X GPU. 
As the main focus of this work is a study of the general multimodal fusion problem, we want to investigate the performance of the two SelectFusion strategies compared to pre-existing models.
In each task, we always choose a deep vision-only model and a deep multimodal model with direct fusion as the \textit{common baselines}. Additionally, specific representative works were chosen as \textit{task baselines}, according to each specific task.
All of our networks including common baselines were trained with a batch size of 16 using the Adam optimizer, with a learning rate $\text{lr}=1e^{-4}$, for a fair comparison. All model were trained for 100 epochs, and the sequence length is chosen as 5.

\subsubsection{Common Baselines}
Common baselines share the same basic framework as our proposed SelectFusion framework. For a fair comparison, the hyper-parameters of proposed network and common baselines are identical, including batch size, learning rate, and the dimension of hidden states. 
The vision-only model is composed of the same visual encoder, temporal modelling and task solver modules as our framework. 
The multimodal model with direct fusion uses the same structure as our proposed framework, except for the fusion component, which is a simple concatenation of the multimodal features.
The single modality model and multimodal model with direct fusion can be viewed as ablated variants of our proposed approach. 
 {In addition, we also compare with a recent multimodal fusion work, i.e. Residual Fusion Network (RFN) \cite{li2021rfn}, which is based on the nest connection incorporated into a residual neural network.
In the task of depth-vision relocalization and LIDAR-vision odometry,  RFN is employed into a framework with the same feature extractors as our select fusion for a fair comparison, but fusion module is replaced with the residual network that aggregates and fuses the features from each extractor. RFN is not employed in visual-inertial odometry, as inertial data are processed with an LSTM, which is not suitable to this CNN based RFN.
}

\subsubsection{Vision-Depth Relocalization}

\noindent \textbf{7-Scenes Dataset (vision+depth):}
The 7-Scenes dataset \cite{Shotton2013} contains RGB images and depth data captured by a handheld Microsoft Kinect camera from seven indoor scenarios. Each scene provides several sequences, and each sequence is with 500-1000 frames of colour and depth images.
We follow the official data split to train and test our models above this dataset.

\noindent \textbf{Task Baselines:}
Our SelectFusion model is built as an end-to-end relocalization model, and thus we compare with LSTM-Pose \cite{walch2017image}, VidLoc \cite{Clark2017}, {and MS-Transformer \cite{shavit2021learning}} which are representative within this category of learning techniques.

\subsubsection{LIDAR-Vision Odometry}

\noindent \textbf{KITTI Odometry Dataset (vision+LIDAR)}
The KITTI Odometry dataset \cite{Geiger2013} provides 11 sequences (00-10) with visual images, LIDAR point cloud and groundtruth. It has been extensively adopted as VO/SLAM benchmark. We use this dataset to fuse the visual and point cloud data to estimate relative pose (odometry) and reconstruct trajectory. Sequences \textit{00, 01, 02, 03, 04, 06, 08, 09} are used for training DNN models, while the rest Sequences \textit{05}, \textit{07}, and \textit{10} are relatively long and used for evaluation. All images are resized to $512 \times 256$.

\noindent \textbf{Task Baselines:}
We {use} three representative works that are evaluated and widely adopted on the KITTI odometry benchmark, as our task baselines, i.e.  VISO2\_M \cite{geiger2011stereoscan}, ORB-SLAM\cite{mur2015orb}, and {ELO \cite{ZhengZ21}}. VISO2\_M is a monocular VO algorithm, in which a fixed camera height, (i.e. a predefined 1.7 meters in the KITTI dataset) is given to recover the absolute scale of generated trajectories. 
{We also adopt ELO \cite{ZhengZ21}, a recent LIDAR odometry work.}

\subsubsection{Visual-Inertial Odometry}

\noindent \textbf{KITTI RAW dataset (visual+inertial)} 
The KITTI Raw dataset \cite{Geiger2013} contains the raw data collection from car-driving scenarios. High-frequency inertial data (100 Hz)  {are} only available in the raw unsynchronized data package. We manually  {synchronize} inertial data and images according to their timestamps, in order to exploit the visual and inertial data to learn odometry estimation. We  {use} Sequences \textit{00, 01, 02, 04, 06, 08, 09} for training and tested the network on Sequences \textit{05}, \textit{07}, and \textit{10}, excluding sequence 03 as the corresponding raw file is unavailable. The images and ground-truth provided by GPS are collected at 10 Hz, while the IMU data  {are} at 100 Hz.

\noindent \textbf{EuRoC MAV dataset (visual+inertial)} 
The EuRoC dataset \cite{euroc} contains tightly synchronized video streams from a Micro Aerial Vehicle (MAV), carrying a stereo camera and an IMU, and is composed by 11 flight trajectories in two environments, exhibiting complex motion. We used Sequence \textit{MH\_05\_difficult} for testing, and left the other sequences for training. We  {downsample} the images and IMUs to 10 Hz and 100 Hz respectively.
\\
\noindent \textbf{Task Baselines:}
We  {choose} four representative VIO pipelines, i.e. MSCKF\cite{Hu2014}, OKVIS\cite{okvis}, mono-VINS\cite{vinsmono} and {VIOLearner \cite{shamwell2019unsupervised}} as task baselines to compare with our deep VIOs: MSCKF\cite{Hu2014} is an Extended Kalman Filter based solution; OKVIS\cite{okvis} is a keyframe based VIO with sliding window nonlinear optimization; mono-VINS\cite{vinsmono} uses sliding window nonlinear optimization and IMU preintegration. VIOLearner {\cite{shamwell2019unsupervised} is a learning based VIO with online error correction. }

            \begin{table*}[t]
    \caption{Vision-depth relocalization (Task 1) on the 7-Scenes dataset, reported in position error (m) and orientation error ($^{\circ}$)}
      \label{tab:7-scenes}
        \small
      \centering
      \begin{tabular}{c|c c c || c c c c}
         Scene  & LSTM-Pose & VidLoc(V+D) & {MS-Transformer} &  {RFN} & Direct Fusion & Soft (Ours) & Hard (Ours) \\
        \hline
        Chess & 0.24 m, 5.77$^{\circ}$  & 0.16 m, NA & {0.11 m, 4.66$^{\circ}$} &  {0.17 m, 5.67$^{\circ}$} & 0.16 m, 5.30$^{\circ}$ & 0.15 m, 5.46$^{\circ}$  & \textbf{0.14 m}, \textbf{5.02$^{\circ}$}\\
            Fire & 0.34 m, 11.9$^{\circ}$  & 0.19 m, NA  & {0.24 m, 9.60$^{\circ}$} &  {0.27 m, 10.3$^{\circ}$} & \textbf{0.26 m}, 10.2$^{\circ}$ & 0.28 m, 10.3$^{\circ}$  & \textbf{0.26 m}, \textbf{9.80$^{\circ}$} \\
            Heads & 0.21 m, 13.7$^{\circ}$  &  0.13 m, NA & {0.14 m, 12.2$^{\circ}$} &  {\textbf{0.14 m}, \textbf{12.0$^{\circ}$}} & 0.16 m, 12.5$^{\circ}$ & {0.15 m}, {12.1$^{\circ}$} & 0.15 m, 12.4$^{\circ}$ \\
            Office  & 0.30 m, 8.08$^{\circ}$   & 0.24 m, NA & {0.17 m, 5.66$^{\circ}$}   &  {0.26 m, 7.22$^{\circ}$} & 0.24 m, 6.78$^{\circ}$ & \textbf{0.22 m}, 6.79$^{\circ}$ & 0.23 m, \textbf{6.39$^{\circ}$}\\
            Pumpkin  & 0.33 m, 7.00$^{\circ}$   & 0.33 m, NA & {0.18 m, 4.44$^{\circ}$}  &  {0.27 m, 5.81$^{\circ}$} & 0.22 m, 5.10$^{\circ}$ & \textbf{0.21 m}, 4.97$^{\circ}$ & \textbf{0.21 m}, \textbf{4.93$^{\circ}$} \\
            Red Kitchen  & 0.37 m, 8.83$^{\circ}$  & 0.28 m, NA & {0.17 m, 5.94$^{\circ}$}  &  {0.31 m, 6.76$^{\circ}$} & \textbf{0.25 m}, 6.41$^{\circ}$ & 0.26 m, \textbf{6.36$^{\circ}$} & \textbf{0.25 m} , 6.76$^{\circ}$ \\ 
            Stairs  & 0.40 m, 13.7$^{\circ}$  &  0.24 m, NA & {0.26 m, 8.45$^{\circ}$}  &  {0.34 m, \textbf{10.8$^{\circ}$}} & 0.37 m, 11.8$^{\circ}$ & 0.35 m, 11.9$^{\circ}$ & \textbf{0.30 m}, {11.3$^{\circ}$} \\
            \hline
            Average & 0.31 m, 9.85$^{\circ}$ & 0.23 m, NA & {0.18 m, 7.28$^{\circ}$}  &  {0.25 m, 8.37$^{\circ}$} & 0.24 m, 8.30$^{\circ}$ & 0.23 m, 8.27$^{\circ}$ & \textbf{0.22 m}, \textbf{8.08$^{\circ}$} \\
      \end{tabular}
      \begin{itemize}
              \footnotesize{
                \item For a fair comparison, the bold character highlights the best results among our proposed approaches and common baselines, excluding task baselines. 
            }
        \end{itemize}
  \end{table*}

\begin{figure}
      \centering
         \includegraphics[width=0.4\textwidth]{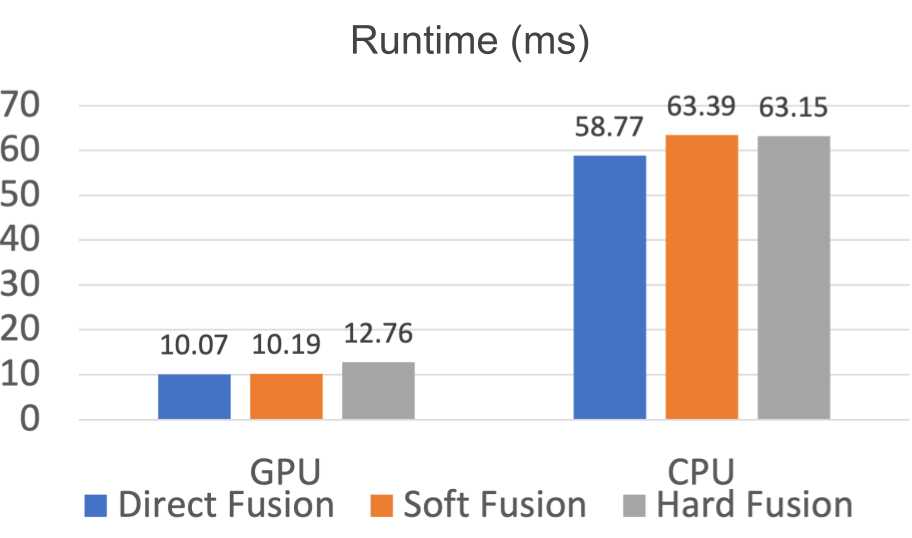}
         \caption{ {The runtime of direct fusion, soft fusion and hard fusion model on a GPU (Geforce RTX 3080Ti) and a CPU (Intel Xeon 2.4G Hz) in the task of learning visual-inertial odometry.}
         }
         \label{fig: running time}
     \end{figure}

 {\subsection{Runtime and Parameters Sensitivity}
This section analyzes the runtime and parameters sensitivity of our proposed fusion mechanisms. 

We first test direct fusion, soft fusion and hard fusion in the task of learning visual-inertial odometry, to collect their prediction time on a NVIDIA RTX 3080Ti GPU and an Intel Xeon 2.4GHz CPU. Fig \ref{fig: running time} reports the averaged results over the per-frame testing time on the Sequence 5 of the KITTI dataset. It is clear to see that no matter our soft fusion or hard fusion only increases the computation burden slightly, compared with direct fusion. For hard fusion model, the runtime of each prediction is 12.76 ms and 63.15 ms on a GPU and CPU respectively. Thus, it can achieve 78 frames per second on a GPU and 15 frames per second on a CPU, which would satisfy the real-time requirement of most robotic applications. 
}

\begin{figure}
      \centering
         \includegraphics[width=0.45\textwidth]{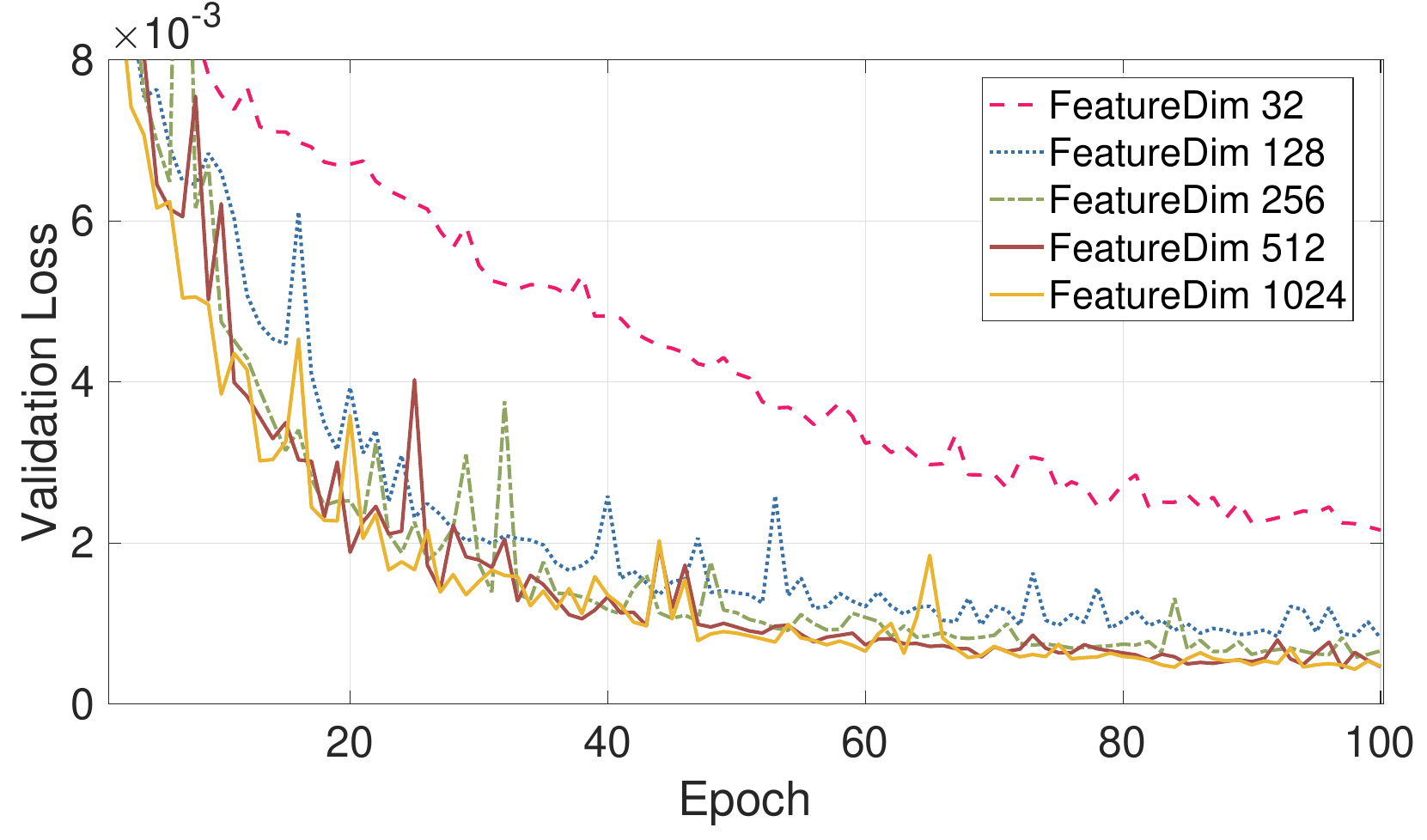}
         \caption{ {The validation losses of proposed hard fusion model in terms of training epochs, with  different feature dimensions of fusion module.}
         }
         \label{fig: feature dim}
     \end{figure}

 {One of the main hyper-parameters inside SelectFusion and baseline frameworks is the feature dimension of fusion module. It determines the dimension of extracted features in the feature extractors and the dimension of  hidden states in the recurrent neural networks. 
To study the influence of this hyper-parameter, we test hard fusion model in the task of visual-inertial odometry with five different feature dimensions from 32 to 1024. 
Fig. \ref{fig: feature dim} shows a comparison of the validation losses in terms of training epochs. Clearly, when increasing the feature dimension from 32 to 128, the validation loss is reduced dramatically. Further, the validation loss decreases slightly, when augmenting the feature dimension to 256, and 512. 
There is no clear change on validation loss if the feature dim is selected as 1024. 
Considering both model performance and memory storage, we thus use 512 as the feature dimension of fusion module in the following experiments.
} 

    \begin{table*}[t]
        \caption{The results of LIDAR-vision odometry (Task 2) on the KITTI Odometry dataset} 
         \label{tab:LIDAR-vision}
        \small
          \centering
          \begin{tabular}{c|c c|| c c c c c c}
              Seq.  & VISO2\_M & {ELO} & Vision Only & LIDAR Only &  {RFN} & Direct Fusion & Soft (Ours) & Hard (Ours) \\
            \hline
        05  & 19.2\%, 17.6$^{\circ}$ & {0.75\%, 0.51$^{\circ}$} & 6.14\%, 2.84$^{\circ}$ & 9.55\%, 3.60$^{\circ}$ &  {5.55\%, 2.22$^{\circ}$}  & 4.73\%, 1.82$^{\circ}$ & 4.65\%, 1.83$^{\circ}$ & \textbf{4.25\%}, \textbf{1.67$^{\circ}$}\\
        07  & 23.6\%, 29.1$^{\circ}$ & {0.60\%, 0.48$^{\circ}$} & 6.41\%, 2.76$^{\circ}$ & 8.63\%, 3.75$^{\circ}$ &  {\textbf{4.19\%}, 1.61$^{\circ}$} & 4.31\%, 2.34$^{\circ}$ & 4.36\%, 2.19$^{\circ}$ & 4.46\%, \textbf{2.17$^{\circ}$}\\
        10 & 41.6\%, 33.0$^{\circ}$ & {2.57\%, 0.84$^{\circ}$} & 6.93\%, 2.97$^{\circ}$ & 15.6\%, 4.77$^{\circ}$ &  {10.3\%, 2.42$^{\circ}$} & 5.92\%, 1.73$^{\circ}$ & 8.35\%, 2.01$^{\circ}$ & \textbf{5.81\%}, \textbf{1.55$^{\circ}$}\\
        \hline
        Ave. & 28.1\%, 26.7$^{\circ}$ & {1.31\%, 0.61$^{\circ}$} & 6.49\%, 2.85$^{\circ}$ & 11.3\%, 4.04$^{\circ}$ &  {6.69\%, 2.08$^{\circ}$}  & 4.99\%, 1.96$^{\circ}$ & 5.78\%, 2.01$^{\circ}$ & \textbf{4.84\%}, \textbf{1.80$^{\circ}$}\\
          \end{tabular}
           \begin{itemize}
              \footnotesize{
                \item $t_{rel}(\%)$ and $r_{rel}(^{\circ})$  are the average translational and rotational RMSE drift (\%) on lengths of 100m-800m.
                \item Vision-Only, LIDAR Only, RFN, Direct Fusion, Soft, and Hard models are trained on Sequence 00, 01, 02, 03, 04, 06, 08 and 09
                \item For a fair comparison, the bold character highlights the best results among our proposed approaches and common baselines, excluding task baselines. 
            }
        \end{itemize}
   \end{table*}
   
   \begin{figure*}
      \centering
         \includegraphics[width=0.85\textwidth]{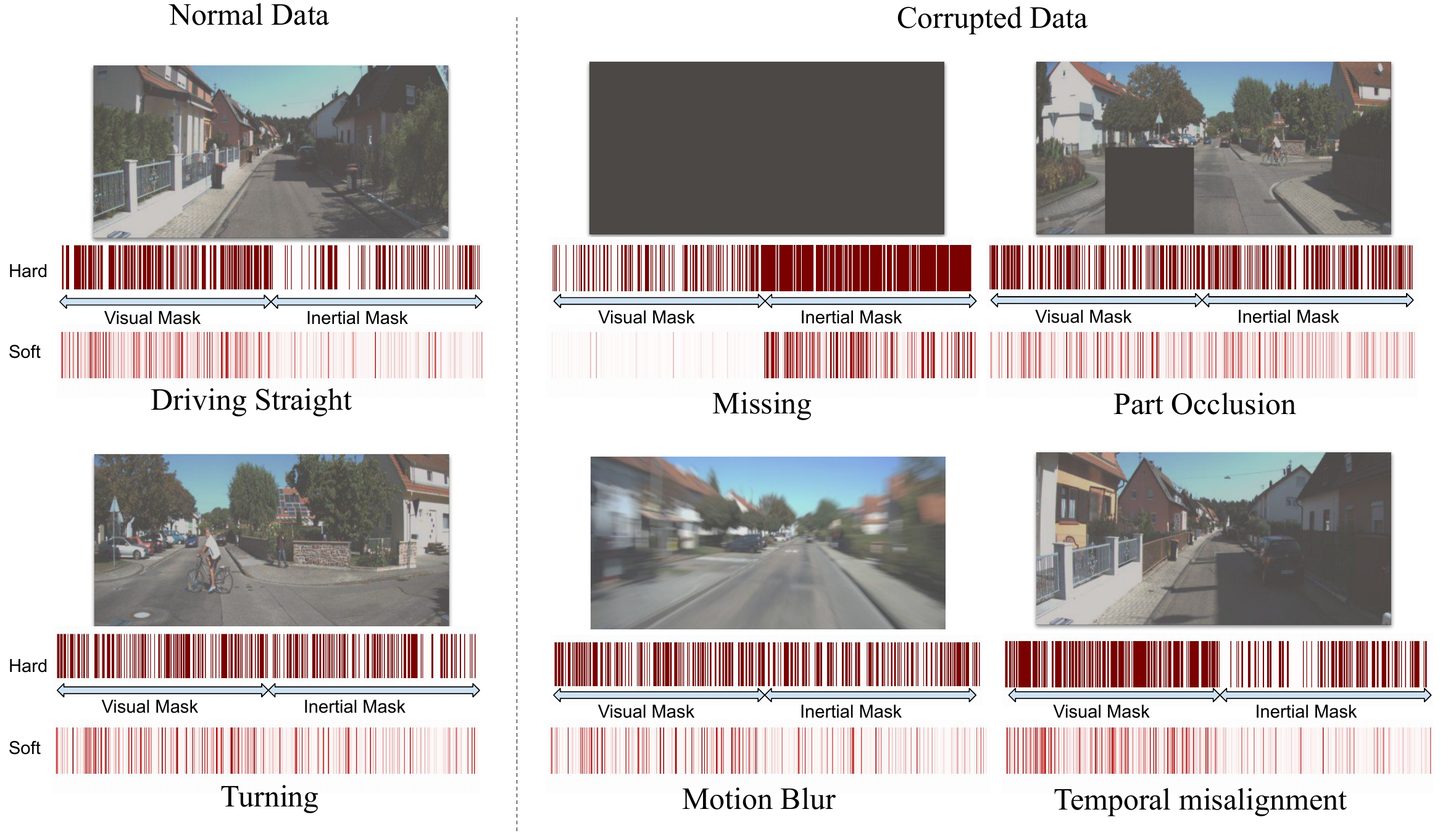}
         \caption{Visualization of the learned hard and soft fusion masks under different conditions for Task 3 Deep VIO on self-driving scenarios (left: normal data; middle and right: corrupted data). The number (hard) or weights (soft) of selected features in the visual and inertial sides can reflect the self-motion dynamics (increasing importance of inertial features during turning), and data corruption conditions. 
         }
         \label{fig:mask}
     \end{figure*}

\subsection{Task 1: Global Relocalization using Vision and Depth}
We first employ selective sensor fusion to combine visual and depth information for a global localization task in indoor scenarios. The features are extracted from RGB and depth images using the visual and depth feature encoders discussed in Section \ref{sec: framework}. Our models, including the common baseline (direct fusion), are trained and evaluated on the 7-Scenes dataset. For each scene, we follow the official data split to train and test our models, and report for each model the median position and orientation error, according to the convention of prior works \cite{kendall2015posenet,walch2017image,Clark2017,Brahmbhatt2018}. 

Table \ref{tab:7-scenes} shows the results for the common baseline (Direct Fusion) and our proposed SelectFusion frameworks, i.e. soft fusion ( Soft (Ours) and hard fusion (Hard (Ours)). For a fair comparison, the only difference in  three models is the feature fusion part.
Clearly, our proposed SelectFusion strategies outperform the common  {baselines}, i.e. direct fusion  {and RFN}. In particular hard fusion further improves the performance of the direct fusion with a gain of 8.33\% in the position and 2.65\% in the orientation. 
 {Although RFN performs best in the Heads scene and achieves most accurate orientation prediction in the Stairs scene, other fusion mechanisms still outperform it in the other scenes and averaged results.
}
This shows the effectiveness of SelectFusion in learning multimodal representation for global relocalization.

In addition to the common  {baselines}, we also choose  {three} representative visual localization approaches as task baselines, i.e. LSTM-Pose\cite{walch2017image}, VidLoc\cite{Clark2017} {and MS-Transformer\cite{shavit2021learning}}. VidLoc can be viewed as a simple direct fusion, but it uses full-size images, and different feature encoders. Our proposed hard fusion model outperforms  LSTM-Pose\cite{walch2017image} and VidLoc\cite{Clark2017}, showing that our models can achieve competitive performance over previous works using only the uniform framework and proposed fusion strategies. {Our method still can not compete with the state-of-the-art  relocalization model, i.e. MS-Transformer with the transformer based architecture. It demonstrates that the performance of our fusion model can be improved with the advances in feature encoders.}

\subsection{Task 2: Deep LIDAR-Vision Odometry}
We now focus on the problem of learning LIDAR-vision odometry in a car-driving scenario. 
The models are trained on the KITTI Odometry dataset and tested on three new sequences, i.e. Sequence 05, 07 and 10. Then the relative poses produced by the neural networks are integrated into global trajectories, which are further evaluated according to the official KITTI VO/SLAM evaluation metrics \cite{Geiger2013}. 
This metric is calculated by averaging the Root Mean Square Errors (RMSEs) of the translation and rotation for all the sub-sequences of lengths (100,..., 800) meters. 

Table \ref{tab:LIDAR-vision} shows the results of our deep LIDAR-vision odometry on the KITTI Odometry dataset. 
Vision Only and LIDAR Only models represent the model using only vision or LIDAR data to estimate ego-motion. Compared with them, fusing vision and LIDAR features (Direct Fusion) contributes to a large improvement no matter in translation or rotation.
Soft (Ours) and Hard (Ours) are our frameworks with soft fusion and hard fusion. 
Our proposed hard fusion is capable of improving the performance over the naive fusion model (i.e.  {the direct fusion}) about 3.0\% in translation and 8.2\% in orientation.  {RFN is able to predict the translation of Sequence 07 accurately, but its overall performance is not as good as other fusion mechanisms in this task.
}
Note that these models are built on the same modules, except the feature fusion part for a fair comparison. 

    \begin{table*}[t]
        \caption{The results of visual-inertial odometry (Task 3) on the KITTI Raw dataset (car-driving scenario)} 
         \label{tab: visual-inertial odometry normal}
        \small
          \centering
          \begin{tabular}{c | c c  c || c c c c}
              Seq.  &  MSCKF & {VINS} & {VIOLearner} & Vision Only & VIO (Direct) & VIO (Soft) & VIO (Hard) \\
\hline
05 &   19.0\%, 82.5$^{\circ}$ & {11.6\%, 1.26$^{\circ}$} & {3.00\%, 1.40$^{\circ}$}  & 6.14\%, 2.84$^{\circ}$ & 4.18\%, 1.57$^{\circ}$ & 4.44\%, 1.69$^{\circ}$ & \textbf{4.11\%}, \textbf{1.49$^{\circ}$}\\
07 &    89.9\%, 126$^{\circ}$ & {10.0\%, 1.72$^{\circ}$} &  {3.60\%, 2.06$^{\circ}$} & 6.41\%, 2.76$^{\circ}$ &  3.39\%, 1.79$^{\circ}$ & \textbf{2.95\%}, \textbf{1.32$^{\circ}$} & 3.44\%, 1.86$^{\circ}$\\
10 &   42.0\%, 134$^{\circ}$ & {16.5\%, 2.34$^{\circ}$} & {2.04\%, 1.37$^{\circ}$} & 6.93\%, 2.97$^{\circ}$ &  2.80\%, 1.69$^{\circ}$ & 2.85\%, 1.22$^{\circ}$ & \textbf{1.51\%}, \textbf{0.91$^{\circ}$}\\
\hline
Ave. &  50.3\%, 114$^{\circ}$ & {12.7\%, 1.77$^{\circ}$} & {2.88\%, 1.61$^{\circ}$} & 6.49\%, 2.85$^{\circ}$ &  3.45\%, 1.69$^{\circ}$ & 3.41\%, \textbf{1.41$^{\circ}$} & \textbf{3.02\%}, 1.42$^{\circ}$\\
          \end{tabular}
           \begin{itemize}
              \footnotesize{
                \item $t_{rel}(\%)$ and $r_{rel}(^{\circ})$ are the average translational (\%) and rotational ($^{\circ}$/100m) RMSE drift on lengths of 100m-800m.
                \item Vision-Only, VIO Direct, VIO Soft, and VIO Hard models are trained on Sequence 00, 01, 02, 04, 06, 08 and 09
                \item For a fair comparison, the bold character highlights the best results among our proposed approaches and common baselines, excluding task baselines. 
            }
        \end{itemize}
   \end{table*}

Meanwhile, three classical methods, i.e. VISO\_M (Monocular Visual Odometry) \cite{geiger2011stereoscan}, {and ELO \cite{ZhengZ21}} are chosen as task baselines to compare with our data-driven approaches. As we can see, the learning based methods greatly outperform the two monocular visual odometry algorithms, but still have a large performance gap with respect to {the traditional LIDAR odometry, i.e. ELO \cite{ZhengZ21}}. The model based methods are tailored to the specific visual odometry or LIDAR odometry problem: ELO is built on scene geometry information and quite accurate with good-quality point cloud data; the monocular visual odometry (VISO\_M) relies on hand-crafted features and it is quite challenging to perform using high-dimensional raw images directly. In comparison, the data-driven models can automatically extract suitable features, which means that they are not restricted to a specific sensor modality or task, hence with the potential to explore an universal framework for deep state estimation.

\subsection{Task 3: Deep VIO on UAV and self-driving scenarios}
Finally, we come to evaluate our proposed model on the KITTI raw dataset (self-driving scenario) and EuRoC MAV dataset (UAV scenario) on learning visual-inertial odometry (VIO). 
These two datasets are challenging, as some real-world sensor degradations are contained in the original data: in the KITTI dataset, some IMU data are missing for a number of timesteps; IMU and camera streams are not tightly time-synchronized, which causes temporal sensor degradation; there are moving vehicles which act to partially occlude the camera; also in the Euroc dataset, there is significant motion blur and camera occlusion. Except the real sensor degradations, we also generate synthetic data degradations above the public datasets to study the robustness of learning models.

\subsubsection{Synthetic Data Degradation}
\label{sec: data degradation}
In order to provide an extensive study of the effects of sensor data degradation and to evaluate the performances of the proposed approach, we generate a degraded dataset, as shown in Figure \ref{fig:mask}, by adding various types of noise and occlusion to the original data, as described in the following.

1) Vision Degradation.

Occlusions: we overlay a mask of dimensions 128$\times$128 pixels on top of the sample images, at random locations for each sample. Occlusions can happen due to dust or dirt on the sensor or stationary objects close to the sensor. 

Motion Blur: we introduce motion blur to represent the camera blur caused by fast ego-motion or fast object movements. This motion blur is generated by estimating the relative motion of the scene, and producing corresponding blur above original images.  
Motion blur can happen when the camera or the light condition changes substantially. 

Missing data: we randomly remove 10\% of the input images. This can occur when packets are dropped from the bus due to excess load or temporary sensor disconnection. It can also occur if we pass through an area of very poor illumination e.g. a tunnel or underpass. 

2) IMU Degradation. 

Noise+bias: on top of the already noisy sensor data we add additive white noise to the accelerometer data and a fixed bias on the gyroscope data. This can occur due to increased sensor temperature and mechanical shocks, causing inevitable thermo-mechanical white noise and random walking noise.

Missing data: we randomly remove windows of inertial samples between two consecutive random visual frames. This can occur when the IMU measuring is unstable or packets are dropped.

 \begin{table}[t]
    \caption{The results (m) of deep visual-inertial odometry (Task 3) on the EuRoC dataset (UAV scenario).}
      \label{tab:euroc}
        \small
      \centering 
      \begin{tabular}{c|c|c|c}
        & Original      & Vision Degrad.          & All Degrad. \\
        \hline
        MSCKF & 0.48 & 30.37 & fail \\
        OKVIS & 0.47 & 1.42 & fail \\
        mono-VINS & 0.35 & fail & fail \\
        \hline
        \hline
        Vision Only & 2.42 & 2.44    & 1.99\\ 
            VIO Direct  & 0.99  & 1.14 & 1.15 \\
            VIO Soft   &  1.06  & 1.18 & 1.21 \\ 
            VIO Hard    &  \textbf{0.84} & \textbf{1.04} & \textbf{1.12} \\ %
      \end{tabular}
      \begin{itemize}
              \footnotesize{
                \item The results (m) are reported the root mean squared error (RMSE) of the absolute translation error (ATE).
                \item Vision-Only, VIO Direct, VIO Soft, and VIO Hard models are trained on the sequences except MH\_05\_difficult of EuRoC MAV dataset \cite{euroc} and tested on Sequence MH\_05\_difficult.
                \item For a fair comparison, the bold character highlights the best results among our approaches and common baselines, excluding task baselines. 
            }
        \end{itemize}
  \end{table}

  \begin{table}[t]
      \caption{The results of deep visual-inertial odometry (Task 3) on the KITII dataset (autonomous driving scenario)}
      \label{tab:kitti}
        \small
      \centering
      \begin{tabular}{c|c|c|c}
         & Original  & Vision Degrad. & All Degrad. \\
        \hline
        Vision Only & 6.49\%, 2.85$^{\circ}$ & 11.8\%, 3.53$^{\circ}$ & 8.06\%, 3.18$^{\circ}$ \\
            VIO Direct & 3.45\%, 1.69$^{\circ}$ & 5.06\%, 1.29$^{\circ}$ & 3.62\%, 1.28$^{\circ}$ \\
            VIO Soft & 3.41\%, \textbf{1.41$^{\circ}$} & \textbf{4.39\%}, 1.84$^{\circ}$ & 3.49\%, 1.40$^{\circ}$ \\
            VIO Hard & \textbf{3.02\%}, 1.42$^{\circ}$ & 4.76\%, \textbf{1.12$^{\circ}$} & \textbf{3.27\%}, \textbf{1.29$^{\circ}$} \\
      \end{tabular}
      \begin{itemize}
              \footnotesize{
                \item $t_{rel}(\%)$ and $r_{rel}(^{\circ})$ are the average translational (\%) and rotational ($^{\circ}$/100m) RMSE drift on lengths of 100m-800m.
                \item Vision-Only, Direct, Soft, and Hard models are trained on Sequence 00, 01, 02, 04, 06, 08 and 09 of KITTI raw dataset \cite{Geiger2013} and tested on Sequence 05, 07 and 10.
                \item For a fair comparison, the bold character highlights the best results among our approaches and common baselines, excluding task baselines. 
            }
        \end{itemize}
  \end{table}

3) Cross-Sensor Degradation.

Spatial misalignment: we randomly alter the relative rotation between the camera and the IMU, compared to the initial extrinsic calibration. This can occur due to axis misalignment and the incorrect sensor calibration.  
We uniformly model up to 10 degrees of misalignment .

Temporal misalignment: we apply a time shift between windows of input images and windows of inertial measurements. This can happen due to relative drifts in clocks between independent sensor subsystems. 

   \begin{table*}[t]
      \caption{Effectiveness of different sensor fusion strategies in presence of different kinds of sensor data corruption for deep VIO} 
      \label{tab:kittiseparatestudy}
        \small
      \centering
      \scalebox{0.99}{
      \begin{tabular}{c|c|c|c|c|c|c|c|}
                      & \multicolumn{3}{c|}{Vision Degradation}
        & \multicolumn{2}{c|}{IMU Degradation} & \multicolumn{2}{c|}{Sensor Degradation} \\
        Model & Occlusion & Blur & Missing & Noise and bias & Missing & Spatial & Temporal \\
        \hline
        Vision Only & 7.23\%, 2.81$^{\circ}$ & 7.76\%, 2.59$^{\circ}$ & 27.6\%, 9.20$^{\circ}$ & 6.49\%, 2.85$^{\circ}$ & 6.49\%, 2.85$^{\circ}$ & 6.49\%, 2.85$^{\circ}$ & 6.49\%, 2.85$^{\circ}$\\
            VIO Direct & 4.24\%, 1.77$^{\circ}$ & 4.28\%, 1.85$^{\circ}$ & 5.61\%, 1.32$^{\circ}$ & 3.74\%, 1.30$^{\circ}$ & 3.59\%, 1.74$^{\circ}$ & 4.12\%, 2.00$^{\circ}$ & 3.27\%, 1.55$^{\circ}$\\
            VIO Soft (Ours) & 3.85\%, \textbf{1.59$^{\circ}$} & 3.82\%, 1.48$^{\circ}$ & 6.42\%, 2.02$^{\circ}$ & 3.72\%, \textbf{1.20$^{\circ}$} & 3.50\%, 1.59$^{\circ}$ & 3.45\%, \textbf{1.46$^{\circ}$} & 3.43\%, 1.72$^{\circ}$\\
            VIO Hard (Ours) & \textbf{3.77\%}, 1.74$^{\circ}$ & \textbf{3.75\%}, \textbf{1.33$^{\circ}$} & \textbf{5.45\%}, \textbf{1.26$^{\circ}$} & \textbf{3.16\%}, 1.64$^{\circ}$ & \textbf{3.18\%}, \textbf{1.47$^{\circ}$} & \textbf{3.22\%}, 1.57$^{\circ}$ & \textbf{3.20\%}, \textbf{1.31$^{\circ}$}\\
      \end{tabular}
      }
      \begin{itemize}
              \footnotesize{
                \item $t_{rel}(\%)$ is the average translational RMSE drift (\%) on lengths of 100m-800m.
                \item $r_{rel}(^{\circ})$ is the average rotational RMSE drift ($^{\circ}$/100m) on lengths of 100m-800m.
                \item The Vision-Only, VIO Direct, VIO Soft, and VIO Hard models are trained on Sequence 00, 01, 02, 04, 06, 08 and 09 of KITTI raw dataset \cite{Geiger2013} with same hyperparameters for a fair comparison, and tested on Sequence 05, 07 and 10.
            }
        \end{itemize}
  \end{table*}
  
\subsubsection{Experiment on the EuRoC Dataset}
In the experiment of EuRoC dataset, we report the root mean squared error (RMSE) of the absolute translation error (ATE) of our models and baselines. This evaluation metric is commonly adopted by previous classical VIO works\cite{Hu2014,okvis,vinsmono}, so that our proposed frameworks can be compared with them directly. 
Table \ref{tab:euroc} reports the performance of learning models (i.e. Vision-Only (DeepVO), VIO Direct (VINet)), and classical algorithms (i.e. MSCKF\cite{Hu2014}, OKVIS\cite{okvis} and mono-VINS\cite{vinsmono}) on the trajectory \textit{MH\_05\_Difficult} in presence of normal data, all combined visual degradation (10\% occlusion, 10\% motion blur, and 10\% missing data) and all combined visual+inertial degradation (5\% for each).
Note that learning models share the same framework and hyper-parameters, while the only difference is the fusion strategy. Details of data degradations can be found at the Section \ref{sec: data degradation}.

We can see that hard fusion consistently outperforms other learning models in all three scenarios, demonstrating the effectiveness of our proposed fusion strategy. In the normal set, hard fusion improves the direct fusion (our common baseline) by 15.15\% in ATE. 
Another notable point is that OKVIS only shows a large performance decrease in the vision degradations, and fails on the all degradations, while mono-VINS fails on both degradation scenarios. In contrast, learning models all can work on degradation scenarios. This indicates that learning models are capable of overcoming sensor degradations to perform more robustly. Learning models still can not compete with the classical algorithms in normal set. This is due to two reasons: 1) in the EuRoC dataset, the groundtruth values of motion tracking (from a Vicon system) and sensor data (from a UAV) are collected from two time systems, and hence the training of learning models is limited because of the probable errors on ground-truth labels; 2) this deep VIO framework is still a basic framework, while extensions and constraints relevant to specific properties of visual and inertial sensors can be added onto it to further enhance the performance, e.g. Bundle adjustment. 
  
  \begin{figure}[t]
      \centering
         \includegraphics[width=0.75\columnwidth]{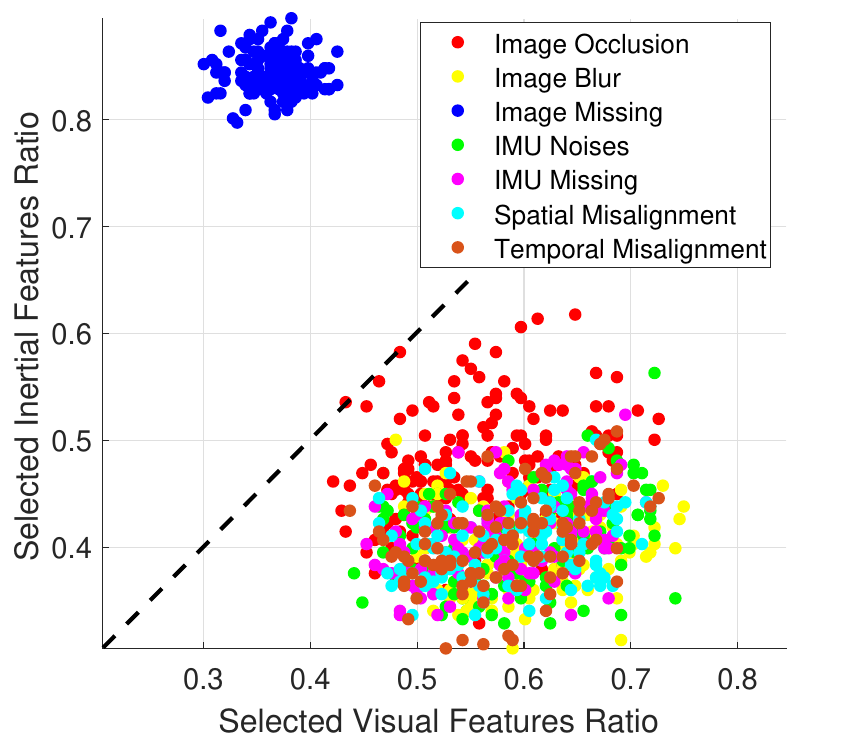}
         \caption{A comparison of visual and inertial features selection rate in seven data degradation scenarios for Task 3.}
         \label{fig:interept_degration}
     \end{figure}
     
         \begin{figure}[t]
      \centering
        \begin{subfigure}[t]{0.22\textwidth}
          \includegraphics[width=\textwidth]{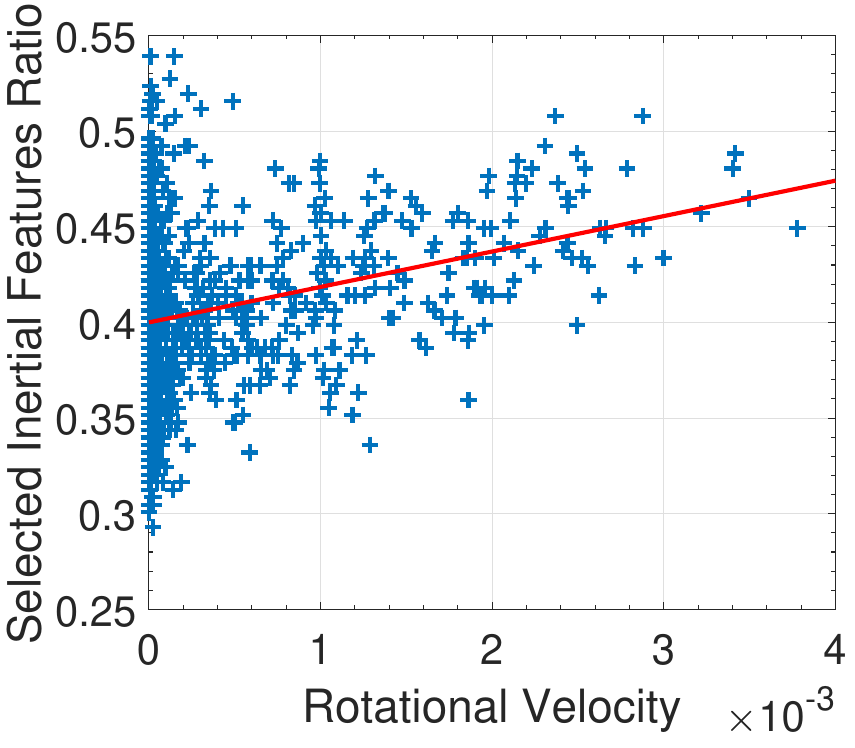}
          \caption{\label{fig:traj_vicon} Inertial-Rotation}
        \end{subfigure}
        \begin{subfigure}[t]{0.22\textwidth}
          \includegraphics[width=\textwidth]{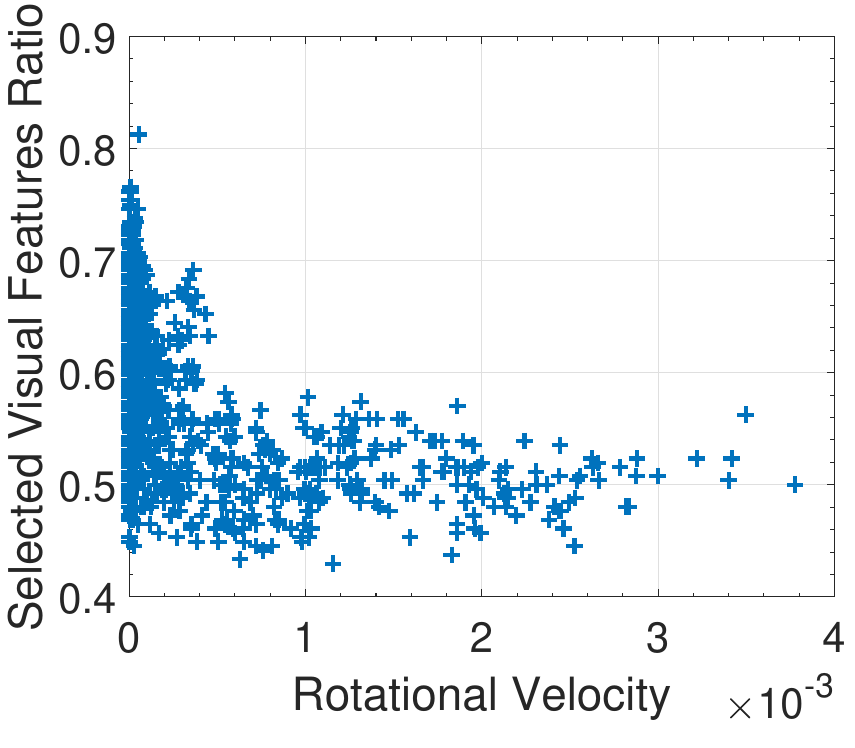}
          \caption{\label{fig:traj_ionet} Visual-Rotation}
        \end{subfigure}
        \begin{subfigure}[t]{0.22\textwidth}
          \includegraphics[width=\textwidth]{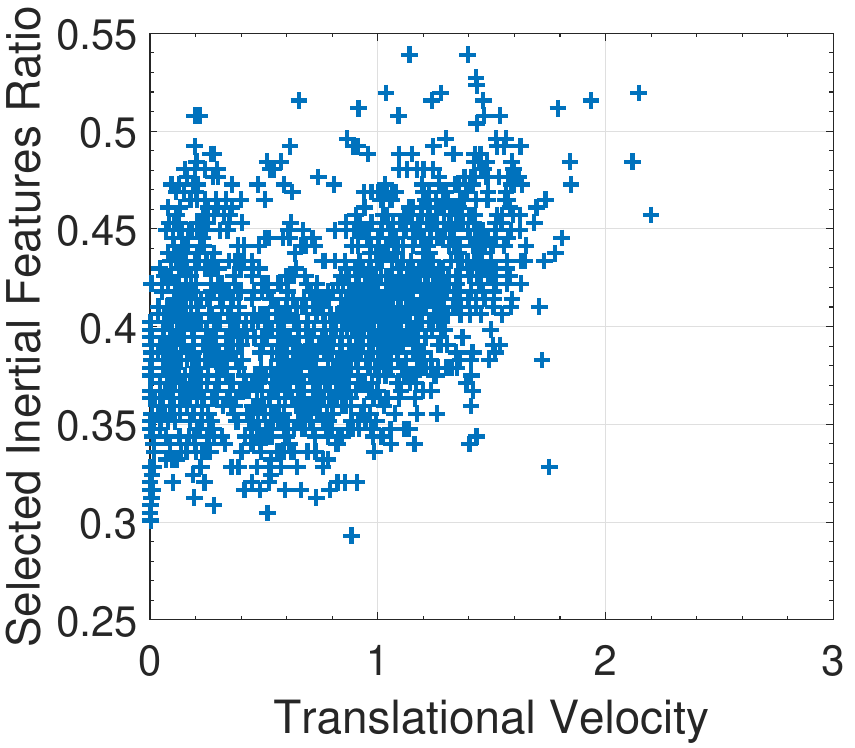}
          \caption{\label{fig:traj_tango} Inertial-Translation}
        \end{subfigure}
         \begin{subfigure}[t]{0.22\textwidth}
          \includegraphics[width=\textwidth]{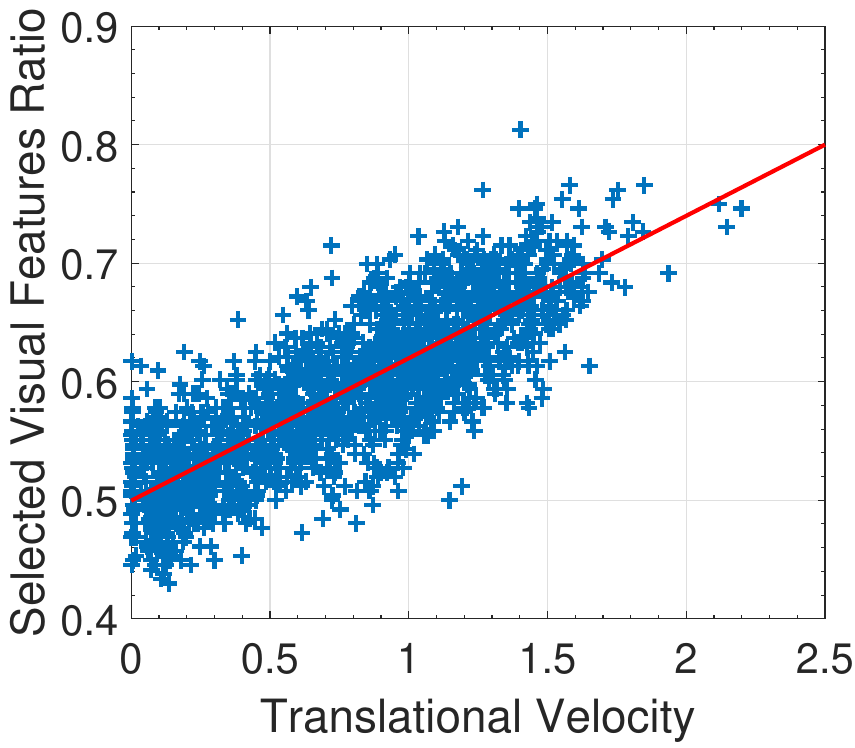}
          \caption{\label{fig:traj_vicon} Visual-Translation}
        \end{subfigure}
        \caption{\label{fig:correlation} Task 3: Correlations between the number of inertial/visual features and amount of rotation/translation show that the inertial features contribute more with rotation rates, e.g. turning, while more visual features are selected with increasing linear velocity.
        }
    \end{figure}

\subsubsection{Experiment on the KITTI Dataset}
On the KITTI dataset, we  {use} the official KITTI VO/SLAM evaluation metrics. 
This metric is to calculate the averaged RMSE of the translation and rotation for all the sub-sequences of lengths (100, ..., 800) meters, which can reflect both the global and local drifts of odometry estimation. 

Table \ref{tab: visual-inertial odometry normal} reports the performance of our proposed hard fusion and soft fusion on the trajectories 05, 07 and 10 of normal dataset, together with {two} classical VIO algorithms (i.e. MSCKF and {VINS}) and {three} learning models, i.e. vision only (DeepVO), VIO direct (VINet) {and VIOlearner}. Our proposed selective sensor fusion (hard) further improves the averaged performance of the direct fusion by 12.46\% in translation and 15.98\% in orientation, while soft fusion shows an improvement of 1.16\% and 16.57\% in translation and rotation. Due to the real issue of bad time synchronization between images and IMUs, OKVIS and mono-VINS failed on the KITTI raw dataset.
We then compare with a MSCKF implementation based on \cite{Hu2014}\footnote{The code can be found at https://uk.mathworks.com/matlabcentral/
fileexchange/43218-visual-inertial-odometry}. This approach, differently from OKVIS or VINS-Mono, is based on a trifocal tensor matching between triplets of successive frames, in order to get an ego-motion estimate, which is then fused with inertial information via a multi-state constraint Kalman filter to refine the estimates of the camera poses for each triplet. This sliding approach arguably makes it more robust to IMU de-synchronisation.
It is clear to see that the learning based VIO models, including VIO (direct), VIO (soft) and VIO (hard) outperform MSCKF and {VINS} by a large margin. This is because MSCKF {and VINS are} limited by the bad time synchronization of two sensors, whilst the learning models are generally more robust to overcome such data degradations caused by real-world issues (data collection). {Note that the original VIOlearner is trained on Sequence 00-08, and tested on Sequence 09 and 10 of the KITTI dataset, and thus it is not fairly to compare with our method directly. But our VIO (hard) still outperforms VIOLearner on Sequence 07 and 10.}

Table \ref{tab:kitti} and \ref{tab:kittiseparatestudy} show the performance of the proposed data fusion strategies, compared with the common baselines on the KITTI raw dataset.
In particular, we compare with a DeepVO \cite{deepvo} (Vision-Only) implementation, and finally with an implementation of VINet \cite{clarkwang2017} (VIO Direct), which uses a na\"ive fusion strategy by concatenating visual and inertial features.
In the vision degraded set the input images are randomly degraded by adding occlusion, motion blurring and removing images, with 10\% probability for each degradation. In the full degradation set, images and IMU sequences from the dataset are corrupted by all seven types of degradation with a probability of 5\% each.
Table \ref{tab:kitti} reports the results of deep VIO models in the presence of combined visual degradations, and all degradations.
As we can see, our proposed selective sensor fusion, especially hard fusion,  {achieves} better performance than the learning models without our proposed fusion module in these sensor degradation scenarios. 
In each data degradation, as illustrated in Table \ref{tab:kittiseparatestudy}, our proposed selective sensor fusion, especially hard fusion consistently outperforms other baselines. This demonstrates to what extent our proposed SelectFusion can tolerant the perturbations of each data degradation, showing that SelectFusion is more robust to the issues caused by data degradations compared with other baselines.
 
\subsection{Interpretation of Selective Fusion} 
Incorporating the hard mask into our framework enables us to quantitatively and qualitatively interpret the fusion process. 
Firstly, we analyse the contribution of each individual modality in different scenarios for deep visual-inertial odometry (Task 3). Since hard fusion blocks some features according to their reliability, in order to interpret the ''feature selection'' mechanism we simply compare the ratio of the non-blocked features for each modality. 
Figure \ref{fig:interept_degration} shows that visual features dominate compared with inertial features in most scenarios. Non-blocked visual features are more than $60\%$, underlining the importance of this modality. 
We see no obvious change when facing small visual degradation, such as image blur, because the FlowNet extractor can deal with such disturbances. However, when the visual degradation becomes stronger the role of inertial features becomes significant. Notably, the two modalities contribute equally in presence of occlusion. As it would be expected, inertial features dominate (by more than $90\%$) with missing images. 

In Figure \ref{fig:correlation} we analyze the correlation between amount of linear and angular velocity and the selected features.
These results also show how the belief on inertial features is stronger in presence of large rotations, e.g. turning, while visual features are more reliable with increasing linear translations. It is interesting to see that at low translational velocity (0.5m / 0.1s) only 50\% to 60\% visual features are activated, while at high speed (1.5m /  0.1s) 60 \% to 75 \% visual features are used. 
     
\section{Conclusion and Future Research}
\label{sec: conclusion}
We present a generic multimodal sensor fusion framework for deep states estimation, in support of odometry estimation and global relocalization tasks. 
Motivated by the need for robust interpretable sensor fusion in real-world applications, we  {propose} two variants of selective fusion modules, i.e. a deterministric soft fusion and a Gumbel-softmax based hard fusion, that can be integrated in different neural network frameworks.
It can learn to perform sensor fusion on feature space from pairs of different modalities, e.g. vision-depth, vision-LIDAR and vision-inertial data, conditioned on the input data itself. Extensive experiments illustrate that our proposed models outperform learning based single modality and multimodal model with direct fusion baselines, and also show competitive performance over other classical approaches, though the performance is still inferior to the domain-specific state-of-the-art in some cases. 
It also demonstrates that learning models are generally more robust than conventional hand-designed algorithms, and our proposed SelectFusion can further improve the performance of basic learning models, and achieve more accurate results than other baselines in these degraded sets.
By interpreting the learned fusion masks, we  {investigate} the influence of different modalities with different types of motion and different levels of sensor degradation. 

\section*{Acknowledgments}
The authors would like to thank Yishu Miao from MoIntelligence, Wei Wu from Tencent and Wei Wang from University of Oxford for helpful discussions. This work was done during Changhao Chen's Ph.D. and postdoctoral study at University of Oxford.

{\small
\bibliographystyle{ieee}
\bibliography{egbib}
}

\begin{IEEEbiography}
[{\includegraphics[width=1in,height=1.25in,clip,keepaspectratio]{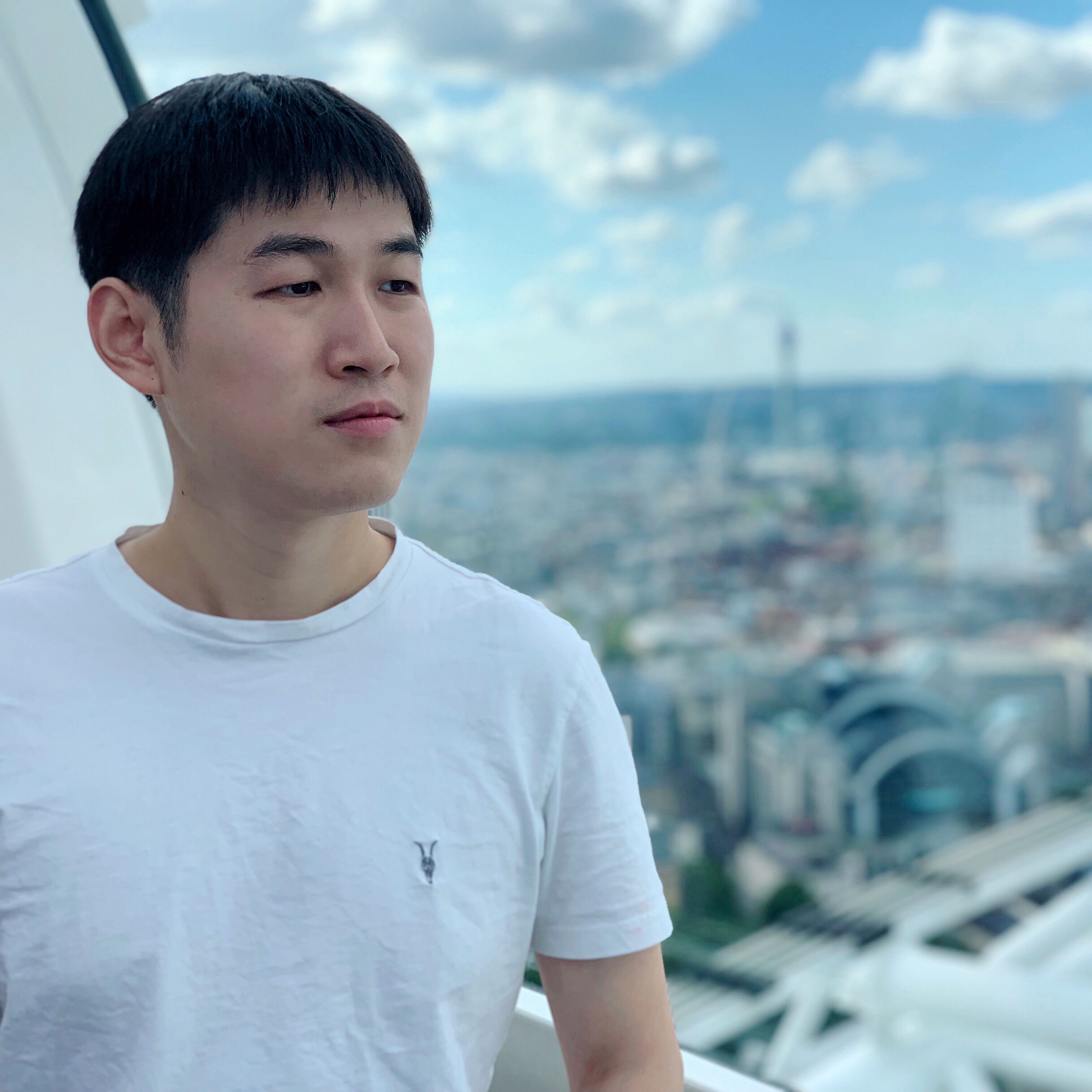}}]{Changhao Chen}
is a Lecturer at College of Intelligence Science and Technology, National University of Defense Technology.
Before that, he obtained his Ph.D. degree at University of Oxford (UK), M.Eng. degree at National University of Defense Technology (China), and B.Eng. degree at Tongji University (China). His research interest lies in robotics, computer vision and cyberphysical systems. 
\end{IEEEbiography}

\begin{IEEEbiography}
[{\includegraphics[width=1in,height=1.25in,clip,keepaspectratio]{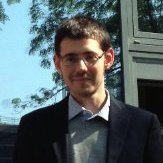}}]{Stefano Rosa}
 is a research fellow at Istituto Italiano di Tecnologia (IIT). He was a postdoctoral researcher at University of Oxford, working on the EPSRC Programme Grant “Mobile Robotics: Enabling a Pervasive Technology of the Future”. He achieved his MS degree in Computer Engineering from Politecnico di Torino in 2008, and his Ph.D. degree in Robotics from Istituto Italiano di Tecnologia (IIT). His research interests include localization and mapping for mobile robotics, computer vision applied to robot navigation, and human-robot interaction.
\end{IEEEbiography}

\begin{IEEEbiography}
[{\includegraphics[width=1in,height=1.25in,clip,keepaspectratio]{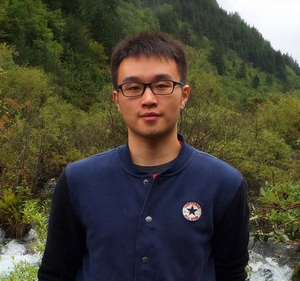}}]{Chris Xiaoxuan Lu}
is an assistant professor at School of Informatics, University of Edinburgh. Before that, he obtained his Ph.D degree at University of Oxford, and MEng degree at Nanyang Technology University,
Singapore. His research interest lies in Cyber Physical Systems, which use networked smart devices to sense and interact with the physical world.
\end{IEEEbiography}

\begin{IEEEbiography}
[{\includegraphics[width=1in,height=1.25in,clip,keepaspectratio]{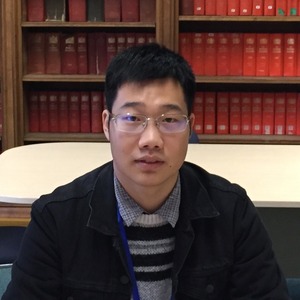}}]{Bing Wang}
 is a Ph.D. student at Department of Computer Science, University of Oxford. Before that, he obtained his BEng Degree at Shenzhen University, China. His research interest lies in camera localization, feature detection, description \& matching, and cross-domain representation learning.
\end{IEEEbiography}

\begin{IEEEbiography}
[{\includegraphics[width=1in,height=1.25in,clip,keepaspectratio]{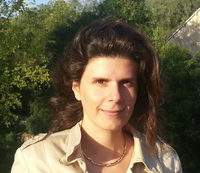}}]{Niki Trigoni}
is a professor at the Department of Computer Science, University of Oxford. She is currently the director of the EPSRC Centre for Doctoral Training on Autonomous Intelligent
Machines and Systems, and leads the Cyber Physical Systems Group. Her research interests lie in intelligent and autonomous sensor systems with applications in positioning, healthcare, environmental monitoring and smart cities.
\end{IEEEbiography}

\begin{IEEEbiography}
[{\includegraphics[width=1in,height=1.25in,clip,keepaspectratio]{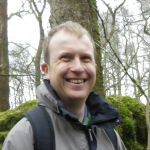}}]{Andrew Markham}
is an associate professor at the Department of Computer Science, University of Oxford. He obtained his BSc (2004)
and PhD (2008) degrees from the University of Cape Town, South Africa. He is the Director of the MSc in Software Engineering. He works on resource-constrained systems, positioning systems,
in particular magneto-inductive positioning and machine intelligence.
\end{IEEEbiography}

\end{document}